\documentclass{report}
\usepackage{setspace}

\usepackage{amssymb,graphicx,color, xcolor, amsmath, latexsym, amsfonts, upgreek}
\usepackage{a4wide, import, comment}
\usepackage{enumitem}
\usepackage{subcaption}

\newtheorem{theorem}{THEOREM}

\newtheorem{definition}[theorem]{DEFINITION}

\usepackage[
backend=biber,
style=numeric,
sorting=nty
]{biblatex}
\title{{ \includegraphics[scale=.5]{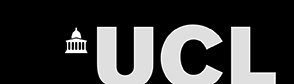}}\\
{{\Huge Catastrophic Forgetting in \\ Continual Reinforcement Learning}} \\
{\large Graphical Structures and the Effects of Task Similarity on Forgetting}
		}
\date{Submission date: 11 September 2023}
\author{Emma Graham \thanks{
{\bf Disclaimer:}
This report is submitted as part requirement for the Machine Learning MSc at UCL. It is
substantially the result of my own work except where explicitly indicated in the text.
The report will be distributed to the internal and external examiners, but thereafter may not be copied or distributed except with permission from the author.}
\\ \\
MSc Machine Learning \\ \\
Andrew Saxe}
\begin{document}
 
\onehalfspacing
\maketitle

{\color{white}{RL}}
\vspace{2 in}

\textbf{\Huge Acknowledgements}

\vspace{0.75 in}

{\centering

Many thanks go to Sebastian Lee, Lee Gunderson, and Grecia Bravo-Hermsdorff for their invaluable expertise, collaboration, and inspiration. I am grateful to Andrew Saxe for the opportunity to work with the Saxe Lab and for introducing me to Sebastian and his project proposal. My appreciation extends to the entire Gatsby Computational Neuroscience Unit for its inclusive support and fostering an environment of collaboration.

}

\newpage 
\begin{abstract}

This paper explores the relationship between task similarity and catastrophic forgetting in reinforcement learning. Catastrophic forgetting, the phenomenon in machine learning of losing the ability to effectively perform on previous tasks, is a significant impediment to continual learning. This study aims to understand the extent to which the similarity of a new task influences the performance on the previous task. Interpretable reinforcement learning, specifically Q-learning, is employed on graph-based tasks with the objective of minimising the number of steps to reach a goal. The study investigates the performance on a previously learned task after training on a new task, for tasks of varying relative levels of complexity. The experimental results reveal a complex dynamic between task similarity and forgetting, with significant fluctuations in forgetting severity observed across degrees of task similarities and task complexities, and are suggestive of an interdependence of forgetting on the similarity and complexity of tasks. The observations were accompanied by observations of high degrees of variability in forgetting and an uneven distribution of task similarity measures. The relationship between these variables remains unclear and no evidence of statistical significance that task similarity has an effect, independently, on forgetting is found in continual reinforcement learning. Further research is warranted to gain a comprehensive understanding of the potential interplay between task similarity and catastrophic forgetting.

\end{abstract} 

\tableofcontents
\setcounter{page}{1}

\chapter{Introduction}

Continual learning, or the ability to sequentially learn tasks, is one of the major unresolved problems in the field of machine learning. A significant barrier to continual learning is the phenomenon known as catastrophic interference or catastrophic forgetting \cite{McCloskey1989CatastrophicProblem, Goodfellow2013MaxoutNetworks}. Catastrophic forgetting occurs when machine learning models lose previously acquired knowledge after learning new information \cite{McCloskey1989CatastrophicProblem, French1999CatastrophicNetworks}.
In reinforcement learning, catastrophic forgetting can substantially degrade the performance on previously learned tasks. Mitigating catastrophic forgetting is needed for efficient learning in complex, real-world environments. Severe forgetting can degrade the performance on previously learned tasks, rendering a reinforcement learning agent ineffective when performing more than one task or adapting between tasks \cite{McCloskey1989CatastrophicProblem, French1999CatastrophicNetworks, Taylor2009TransferSurvey}.
The relationship between forgetting and task similarity remains an open research problem that has not yet been quantified in the reinforcement learning setting. The aim of this research is to quantify the effects of the new task difference, on the performance of the initial task in continual reinforcement learning. This chapter will overview background topics, define the research problem and objectives, and layout the scope and structure of this study.


\section{Background Topics}

Intelligence will be defined as the ability to learn to make decisions to achieve goals. While biological intelligence is the intellect of living systems, artificial intelligence is the intellectual ability imbued in non-living systems. 

Machine learning is the term for computer systems that are able to learn and adapt without following explicit instructions, essentially able to teach themselves. This is a subset of artificial intelligence and is segmented into three general categories: supervised learning, unsupervised learning, and reinforcement learning. Broadly speaking, reinforcement learning is a sequential decision problem comprising goal-directed learning from interaction \cite{Sutton2018Reinforcement2018, Jaimungal2022ReinforcementOptimisation}. 

\begin{figure}[ht]
\centering
\includegraphics[width=0.5\textwidth]{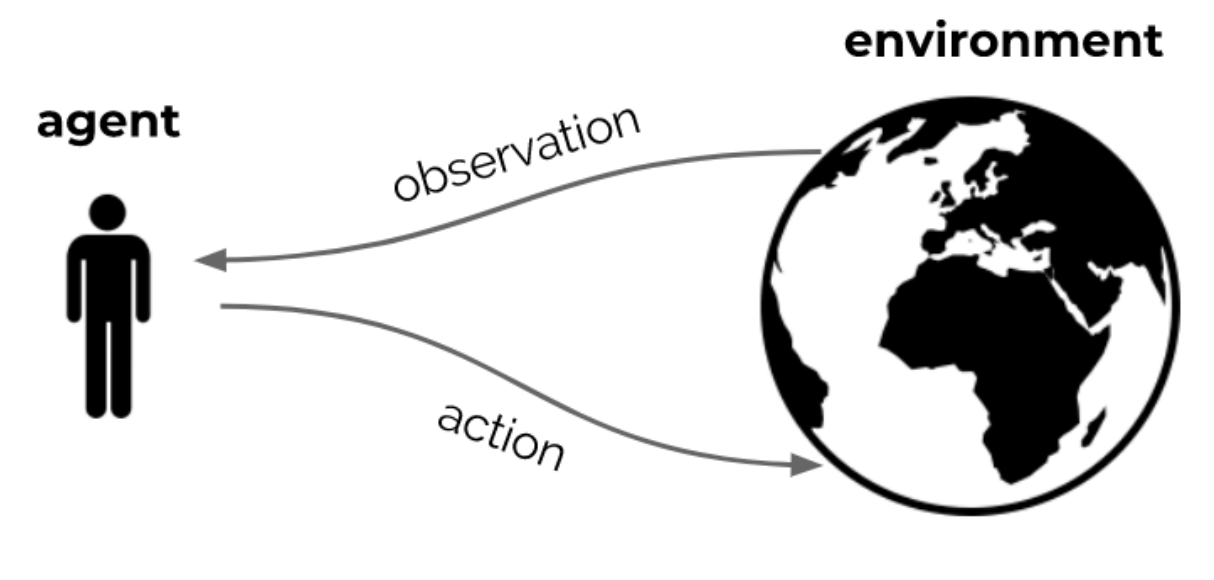}
\caption{Supporting graphic depicting reinforcement learning from the Reinforcement Learning taught class (COMP0089) by DeepMind at UCL \cite{HadovanHasselt2023ReinforcementLearning}.}
\label{fig:RL}
\end{figure}

In reinforcement learning, the inherent interaction with the environment, as depicted in the Figure \ref{fig:RL}, is a powerful representation of the dynamic, continuous feedback loop that characterises our own learning paradigm. Notably, the reinforcement learning process has been shown to mirror the neuromodulatory influences that drive attention and decision-making in biological systems \cite{Yu2005UncertaintyAttention}. This positions reinforcement learning not merely as a methodology within machine learning, but as a paradigm that bridges artificial and biological learning processes \cite{Mnih2015Human-levelLearning, Schultz1997AReward}.

The act of reaching the goal(s) of a reinforcement learning agent is frequently called the task. The concept of a task is fundamental to machine learning because it helps define the problem that the machine learning system is trying to solve. A classical definition of a task in machine learning is: "A task T is a measure of performance that the machine learning system should improve with experience E on some class of tasks T, given some performance measure P" \cite{Mitchell1997Machine1997}. Task similarity in machine learning refers to the degree to which two tasks share underlying patterns, structures, or concepts, indicating the potential for shared or overlapping learning between these tasks \cite{Zhang2020SurveyLearning}.

A graph is a concept from mathematics that aids in the understanding and visualisation of the relationships among different objects. Graph theory is a prominent field within discrete mathematics, which studies the properties of graph structures and will be overviewed and formalised in the Graph Theory section of Chapter 2 \cite{West2012IntroductionEdition}.

Computational Geometry is a branch of computer science that deals with the mathematical techniques necessary for the manipulation and analysis of geometric structures and shapes. This field is especially important in areas where solving problems involving the spatial and structural relationships between different objects is needed. The Computational Geometry section of Chapter 2 will outline some relevant fundamental concepts of this discipline.


\section{Research Problem}

Catastrophic forgetting poses a significant hurdle in machine learning, causing detrimental effects on the performance of learning models. Understanding its impact and the factors that influence it, such as task similarity, will help enable the development of robust and efficient learning algorithms. Understanding the relationship between task differences and memory retention is a key component for the development of effective continual learning models in reinforcement learning. However, this relationship has been studied less than the amount of knowledge transferred to new tasks.

These insights will aid in mitigating the effects of catastrophic forgetting and enhance our understanding of continual learning. The ability of an artificially intelligent system to learn new tasks without forgetting previous ones would be a remarkable capability. This skill is a powerful aspect of intelligence, both natural and artificial, and forms the basis for adaptability, versatility, and efficiency.

\section{Research Objectives}

The objective of this study is to understand the relationship between catastrophic forgetting and task similarity in continual reinforcement learning. The research seeks to offer a new understanding of how the similarity of a new task impacts the performance on the previous task in an interpretable reinforcement learning setting. To elucidate this relationship, the study aims to produce a graph depicting the interplay between forgetting and task similarity. Consequentially, methods to quantify both measures, that of the amount forgotten and the similarity of the task, are essential.


\section{Scope}

This study uses tabular reinforcement learning, specifically utilising Q-learning. Also, the research is restricted to graph-based reinforcement learning with the objective of minimising the quantity of steps to reach a goal. The study delves into the effects on the performance of a previously learned task after a single training session on a new task. The relationship between forgetting and task similarity is examined in tasks of varying relative levels of complexity, of which a modern laptop has the capacity to capture.


\section{Structural Outline}

This report details background knowledge of the concepts used in this study in Chapter 2. Starting with an overview of reinforcement learning, this section lays out fundamentals of continual learning and catastrophic forgetting. Task similarity and pertinent principles of graph theory and computational geometry follow in detail. The next chapter, Chapter 3, gives a literature review of current research on catastrophic forgetting, task similarity, and their relationship. The methodology of the study is expressed in Chapter 4. In this chapter the reinforcement learning models and the methods for creating similar tasks, measuring the difference between tasks, and quantifying the amount of forgetting are outlined and detailed. The rationale behind the modelling choices and quantification metrics are discussed throughout this section. Chapter 5 is the results section which presents the findings from a hundred trials. The tasks generated, the similarity measures of the tasks, and the difference in performance quantities are analysed. Then, the relationship between the task similarity and the amount forgotten is discussed and limitations to the study identified. In Chapter 6, the paper finishes with the conclusions, their implications in reinforcement learning and continual learning, and directions for future work.
\chapter{Background}


\section{Reinforcement Learning}

Reinforcement Learning (RL) is a type of machine learning where an agent learns to make decisions by interacting with an environment \cite{Sutton2018Reinforcement2018}. RL is goal-directed and makes decisions as to obtain its objectives. Reinforcement learning is based on the reward hypothesis, the hypothesis being that any goal can be formalised as the outcome of maximising a cumulative reward. The agent tries to maximise a cumulative reward by learning a policy for choosing actions based on the environmental states. 

\subsection{Reinforcement Learning Formalism}

The RL learning formalism includes the environment represents the learning dynamics of the problem, the reward signal which specifies the goal, and the agent containing the agent state, the policy, the value function, and a model. In RL, an agent's environment and reward signal are typically modelled using a Markov Decision Process (MDP).

\begin{definition}
    A \textbf{Markov Decision Process} is a tuple $\langle \mathcal{S}, \mathcal{A}, \mathcal{P}, \mathcal{R}, \gamma \rangle$, where:

\begin{itemize}[noitemsep]
    \item $\mathcal{S}$ is a finite set of states.
    \item $\mathcal{A}$ is a finite set of actions.
    \item $\mathcal{P} : \mathcal{S} \times \mathcal{A} \times \mathcal{S} \rightarrow [0, 1]$ are the state transition probability functions, where $P(s, a, s')$ is the probability of reaching state $s'$ when taking action $a$ in state $s$.
    \item $\mathcal{R} : \mathcal{S} \times \mathcal{A} \rightarrow \mathbb{R}$ are the reward functions, where $R(s, a)$ is the expected immediate reward for taking action $a$ in state $s$.
    \item $\gamma \in [0, 1]$ is the discount factor, which determines the importance of future rewards.
\end{itemize}
\end{definition}

The agent state is the internal state of the agent as opposed to the state of the environment the agent is in. The policy is the map from agent state to action, which defines the agent's behaviour. The goal of the agent in a reinforcement learning problem is to learn a policy $\pi: \mathcal{S} \rightarrow \mathcal{A}$, a mapping from states to actions, to maximise the expected sum of discounted rewards called the return, denoted by $G_t$ in equation (2.1).

\begin{equation}
    G_t = \mathbb{E} \left[ \sum_{k=0}^{\infty} \gamma^k R_{s_t, a_t} \right]
\end{equation}

\subsection{Approaches to Learning the Optimal Policy}

To learn the optimal policy, the agent often represents the value of a state or a state-action pair through value functions. The value function estimate is the expected return. The state-value function $V^\pi(s)$ is the expected return when starting from state $s$ and following policy $\pi$, while the action-value function $Q^\pi(s, a)$ is the expected return when starting from state $s$, taking action $a$, and then following policy $\pi$ \cite{Sutton2018Reinforcement2018}:

\begin{equation}
    V^\pi(s) = \mathbb{E}_\pi[G_t | S_t = s],
\end{equation}

\begin{equation}
    Q^\pi(s, a) = \mathbb{E}_\pi[G_t | S_t = s, A_t = a].
\end{equation}

A model predicts what the environment will do next, for instance the expected next state or reward if a certain action is taken in the current state.

In reinforcement learning, there are two fundamental approaches to learning the optimal policy $\pi^*$: value-based methods and policy-gradient methods. Both of these methods can be applied in a variety of settings, including both tabular and function approximation scenarios.

\subsubsection{Value-Based Methods}

Value-based methods aim to learn the value function and indirectly derive the policy from this value function. A core concept in these methods is Temporal-Difference learning.

Temporal-Difference learning (TD) learns a state-value function $V(s)$ which takes in a state and outputs the expected cumulative reward following a certain policy. So, TD learning methods learn directly from raw experience without requiring a model of the environment's dynamics. This is accomplished through a process called bootstrapping, where estimates are updated based on other estimates. The simplest form of TD learning is TD(0), or one-step TD, which estimates the value function $V(s)$ of a policy. The update rule for TD(0) is:

\begin{equation}
    V(s_t) \gets V(s_t) + \alpha [R_{t+1} + \gamma V(s_{t+1}) - V(s_t)]
\end{equation}

where $s_t$ is the current state, $R_{t+1}$ is the reward received after transitioning from state $s_t$ to state $s_{t+1}$, $V(s_{t+1})$ is the estimate of the value of state $s_{t+1}$, $\alpha$ is the learning rate, and $\gamma$ is the discount factor \cite{Sutton2018Reinforcement2018}.

A common value-based method is Q-learning, which is a type of TD learning that learns the action-value function, or Q-function. Q-Learning updates an action-value function, called the Q-function, which takes a state-action pair as input and outputs the expected cumulative reward for taking that action in that state, following a certain policy \cite{Watkins1992Q-learning}. The update rule for Q-learning is:

\begin{equation}
    Q(s_t, a_t) \gets Q(s_t, a_t) + \alpha [R_{t+1} + \gamma \max_{a'} Q(s_{t+1}, a') - Q(s_t, a_t)]
\end{equation}

where $s_t$ is the current state, $a_t$ is the current action, $R_{t+1}$ is the reward received after taking action $a_t$ in state $s_t$, $s_{t+1}$ is the new state, $a'$ refers to any possible action in state $s_{t+1}$, $\alpha$ is the learning rate, and $\gamma$ is the discount factor. 

\subsubsection{Policy-Gradient Methods}

Policy-gradient methods learn a policy that maximises the expected return by directly optimising the policy parameters. These methods are based on the idea of using gradient ascent to find the optimal parameters that maximise the expected return.

One common policy gradient method is the REINFORCE algorithm. The REINFORCE algorithm works by adjusting the policy parameters using the following update rule:

\begin{equation}
\theta \gets \theta + \alpha \nabla_\theta \: log \: \pi_\theta(a_t|s_t) G_t
\end{equation}

where $\theta$ are the policy parameters, $\alpha$ is the learning rate, $\pi_\theta(a_t|s_t)$ is the probability of taking action $a_t$ in state $s_t$ under policy $\pi$, and $G_t$ is the return from time step $t$ \cite{Willia1992SimpleLearning}.

Another policy-gradient approach is the Actor-Critic method, which combines the benefits of policy-gradient methods and value-based methods. In Actor-Critic methods, two separate neural networks are employed: the Actor, which updates the policy, and the Critic, which evaluates the policy by computing the value function. This allows the algorithm to leverage the strengths of both types of methods. The Critic reduces the variance of the gradient estimation, which can help improve the stability and speed of learning.

The update rule for the Actor-Critic method is:

\begin{equation}
\theta \gets \theta + \alpha \nabla_\theta \: log \: \pi_\theta(a_t|s_t) V_{\phi}(s_t)
\end{equation}

where $\theta$ and $\phi$ are the policy parameters and value function parameters, respectively, $\alpha$ is the learning rate, $\pi_\theta(a_t|s_t)$ is the probability of taking action $a_t$ in state $s_t$ under policy $\pi$, and $V_{\phi}(s_t)$ is the value function estimated by the Critic at state $s_t$ \cite{Konda2003OnAlgorithms}.

\subsection{Tabular Reinforcement Learning}

Tabular Reinforcement Learning is a class of reinforcement learning algorithms that operate in discrete and finite state-action spaces \cite{Sutton2018Reinforcement2018}. In these settings, the agent learns the optimal policy by interacting with the environment using a table containing values or policies for each state-action pair. The agent maintains and updates a table for $V^\pi(s)$ or $Q^\pi(s, a)$ as it interacts with the environment. 

Tabular reinforcement learning has the advantage of being understandable and offering strong theoretical guarantees. These methods are limited to problems with small state and action spaces due to the need to maintain and update a table of values for each state-action pair. For problems where the state and action spaces are too large to feasibly create such a table, function approximation methods, including both value-based and policy-gradient methods, are used to approximate the value function or policy function with a parameterised function, such as a neural network.

\subsection{Action Selection Policies}

In RL, exploration policies are used to balance the trade-off between exploration (discovering new information about the environment) and exploitation (using the current knowledge to maximise rewards). Several popular exploration policies are the $\epsilon$-greedy policy, the Boltzmann exploration, and the Upper Confidence Bound method. \\

I. $\epsilon$-greedy policy: This is one of the simplest and most widely used exploration policies in RL \cite{Sutton2018Reinforcement2018}. In the $\epsilon$-greedy policy, the agent selects a random action with probability $\epsilon$ and chooses the action that currently has the highest estimated value with probability $1 - \epsilon$. This policy is defined in Definition 2.

\begin{definition} 
The \textbf{$\epsilon$-greedy policy}, $\pi(a|s)$ selects an action $a$ given the state $s$ s.t. 
$$
\pi(a|s) =
\left\{
    \begin{array}{lr}
        1 - \epsilon + \frac{\epsilon}{|\mathcal{A}|} & \text{if } a = \arg\max_{a'} Q(s, a')\\
        \frac{\epsilon}{|\mathcal{A}|}  & \text{otherwise}
    \end{array}
\right\}
$$
where $Q(s, a)$ is the action-value function, and $|\mathcal{A}|$ is the size of the action space.
\end{definition}

II. Boltzmann exploration or Softmax policy: This policy selects actions based on their relative action-value estimates rather than selecting the action with the highest value deterministically \cite{Sutton2018Reinforcement2018}. The function works, essentially, as a smoothing function that transforms a vector of real numbers into a probability distribution, with each probability representing the relative preference of the agent for each action. The mathematical representation of the softmax action selection is given in Definition 3.

\begin{definition}
    The \textbf{softmax policy} $\pi$ computes the probability of selecting an action $a$ given a state $s$ by
    \\
    $$\pi(a|s) = \frac{\exp(Q(s, a) / \tau)}{\sum_{a'} \exp(Q(s, a') / \tau)}$$
    \\
    with a temperature parameter $\tau > 0$.
\end{definition}

The temperature parameter $\tau$ controls the exploration-exploitation trade-off. Higher values of $\tau$ result in more exploration, while lower values make the policy more deterministic and focused on exploitation. \\

III. Upper Confidence Bound (UCB) action selection: UCB is an exploration policy that balances exploration and exploitation by considering the uncertainty in the action-value estimates \cite{Auer2002Finite-timeProblem}. The UCB action selection rule is defined in Definition 4.

\begin{definition}
    The \textbf{Upper Confidence Bound action selection policy} is the selection of the action $a_t$ given state $s$ by \\ 
    $$a_t = \arg\max_{a} \left[ Q(s, a) + c \sqrt{\frac{\log t}{N_t(a)}} \right]$$ 
    \\
    where $t$ is the current time step, $N_t(a)$ is the number of times action $a$ has been selected up to time $t$, and $c > 0$ is a parameter that controls the exploration-exploitation trade-off. The second term in the square brackets represents the uncertainty in the action-value estimates and encourages exploration of actions that have been tried less frequently.
\end{definition}

\section{Continual Learning}

Continual learning is a concept that refers to the ability of a model to learn tasks sequentially. This ability to keep learning, without forgetting previously learned information, is one of the major unresolved challenges in the field. In a continual learning framework, a model is presented with a series of tasks $T_1, T_2, \ldots, T_n$. The model learns task $T_i$ at time $i$, while retaining knowledge from tasks $T_1, \ldots, T_{i-1}$, and without access to data from future tasks $T_{i+1}, \ldots, T_n$ \cite{Lopez-Paz2017GradientLearning}.

\subsection{Forgetting and Transfer Learning}

The ability of a model to apply its knowledge to new tasks, a process known as transfer learning, is a key aspect of continual learning. Similarly, the ability to maintain performance on original tasks, or the lack of \textit{forgetting}, is another crucial component. Together, transfer learning and forgetting represent the central challenge in continual learning: how to effectively transfer knowledge to new tasks without forgetting important information from previous ones. This balance is often referred to as the stability-plasticity dilemma in continual learning.

\subsubsection{Continual Supervised Learning}

In the realm of continual supervised learning, each task $T_i$ typically involves a classification or regression challenge with its own training dataset $D_i = {(x_{i,j}, y_{i,j})}{j=1}^{N_i}$, where $x{i,j}$ is the $j$-th input example, $y_{i,j}$ is the corresponding label, and $N_i$ is the number of examples for task $T_i$ \cite{Rebuffi2016ICaRL:Learning}.

\subsubsection{Continual Unsupervised Learning}

Continual unsupervised learning involves tasks $T_i$ that require understanding the underlying data distribution or identifying other properties of the data, without the use of explicit labels. A task could involve clustering a different dataset or conducting a dimensionality reduction task with varying objectives \cite{Aljundi2019OnlineRetrieval}.

\subsubsection{Continual Reinforcement Learning}

The framework of reinforcement learning can simply be defined for each task when in the continual learning context. 
\newpage 
\begin{definition}
    Continual reinforcement learning tasks $T_i$ are defined as a Markov decision process (MDP) in a tuple represented by a tuple $\langle \mathcal{S}_i, \mathcal{A}_i, \mathcal{P}_i, \mathcal{R}_i, \gamma_i \rangle$ :

    \begin{itemize}[noitemsep]
        \item $\mathcal{S}_i$ is a finite set of states for task $T_i$.
        \item $\mathcal{A}_i$ is a finite set of actions for task $T_i$.
        \item $\mathcal{P}_i : \mathcal{S}_i \times \mathcal{A}_i \times \mathcal{S}_i \rightarrow [0, 1]$ represents the state transition probability functions for task $T_i$, where $\mathcal{P}_i(s, a, s')$ is the probability of reaching state $s'$ when taking action $a$ in state $s$.
        \item $\mathcal{R}_i : \mathcal{S}_i \times \mathcal{A}_i \rightarrow \mathbb{R}$ represents the reward functions for task $T_i$, where $\mathcal{R}_i(s, a)$ is the expected immediate reward for taking action $a$ in state $s$.
        \item $\gamma_i \in [0, 1]$ is the discount factor for task $T_i$, which determines the importance of future rewards.
    \end{itemize}
\end{definition}

This definition aligns with the general Markov decision process, but in the context of continual learning, each task has its own MDP \cite{Ring1994ContinualEnvironments}.

\subsection{Catastrophic Forgetting}

A significant barrier to continual learning is the phenomenon known as catastrophic interference or catastrophic forgetting \cite{McCloskey1989CatastrophicProblem, Goodfellow2013MaxoutNetworks}. Catastrophic forgetting refers to the loss of previously acquired knowledge by ML models when learning new information \cite{McCloskey1989CatastrophicProblem, French1999CatastrophicNetworks}. This issue can arise when weights, essential for the initial task, are overwritten by information pertinent to the new task, causing the model to struggle to perform on the initial task after learning another.

\section{Task Similarity}

Task similarity, the quantification of difference between two tasks, can be quite challenging in the domain of RL. Within the conventional Markov Decision Process (MDP) framework, various metrics are proposed to assess task similarity \cite{Garcia2022AProcesses}. These metrics broadly classify into two categories: model-based metrics and performance-based metrics.

Model-based metrics provide an estimation of the similarity between an original and a new task by using the respective MDP model parameters of each. On the contrary, performance-based metrics are dependent on the performance of the agents in the base and new tasks.

Types of model-based metrics are dependent on which elements of the MDPs are considered and can be categorised into the following:

\begin{itemize}
    \item \textit{Transition and reward dynamics:} Metrics in this category necessitate comprehensive knowledge of the MDP models for both the original and new tasks. Three main approaches are identified: state abstraction techniques \cite{Li2006TowardsMDPs, Ferns2004MetricsProcesses, Ferns2012MetricsSpaces, Castro2020ScalableProcesses}, compliance metrics \cite{Lazaric2008TransferLearning, Fachantidis2015TransferSelection, AnestisFachantidis2016KnowledgeLearning.}, and metrics that construct MDP graphs \cite{Kuhlmann2007Graph-basedGames, Wang2019MeasuringMDPs}.
    \item \textit{Transitions:} Metrics in this category use tuples in the form $\langle s,a,s' \rangle$ to measure the similarity between MDPs \cite{Taylor2008AutonomousLearning, Ammar2014AnLearning}. 
    \item \textit{Rewards:} Metrics in this category use tuples in the form $\langle s,a,r \rangle$ to measure the similarity between MDPs \cite{Carroll2005TaskLibraries, Tao2021REPAINT:Learning, Gleave2021QUANTIFYINGFUNCTIONS}.
    \item \textit{State and actions:} Metrics in this category use pairs in the form $\langle s,a \rangle$ in both the source task and the target task to compute the similarity between MDPs \cite{Carroll2005TaskLibraries, Taylor2009BoundingHomomorphisms, Narayan2019EffectsLearning}.
    \item \textit{States:} Metrics in this category use the state space in both the original and new tasks to compute their similarity \cite{Svetlik2017AutomaticAgents}. Another relevant area of research in this category is case-based reasoning, where RL approaches use a similarity function between the states in the new task and the states in a case base from a previous source task \cite{Agnar1994Case-BasedApproaches, Celiberto2011UsingApplication}.
\end{itemize}

For performance-based metrics, two distinct approaches are identified: policy similarity and transfer gain.

\begin{enumerate}
    \item \textit{Policy similarity:} Metrics in this category are based on the learned value function $V^\pi$ or the action-value function $Q^\pi$, or the behavioural policies $\pi$ obtained in the base task and the new task.
    \item \textit{Transfer gain:} Approaches in this category approximate the level of similarity to the advantage gained by using the knowledge in one original task to speed the learning of another task \cite{Carroll2005TaskLibraries, Taylor2009TransferSurvey}.
\end{enumerate}

Each of these metrics measure a different component in the RL framework. Assigning a comprehensive metric to similarity can be quite difficult. Additionally, the concept of similarity can be subjective, depending on the specific problem domain and the goals of the learning process \cite{Taylor2009TransferSurvey}.

\subsubsection{Simple Example of the Subjectivity of Similarity Quantification}

Consider evaluating the 'similarity' of three types of vehicles: sports cars, sedans, and pickup trucks. Both considered passenger cars, perhaps sports cars have more in common with sedans. To quantify this relationship, the 'similarity score' of sedans from sports cars could be 10. Considered a utility vehicle and not a passenger car, a pickup truck could receive a lower 'similarity score', but how much lower? How different does the categorisation make pickup truck's 'similarity score' and how different relative to the 'similarity score' of the sedan? 
However, when the 'similarity' evaluation shifts to vehicle performance capabilities, a contrasting scenario can emerge. It could be seen that sports cars and pickup trucks, both exhibiting high speed and power, are less different than sports cars and sedans, contradicting the previous numerical measure. This example demonstrates how the quantification of 'similarity' can be significantly influenced by the particular characteristics or metrics selected for the comparison process.

\section{Graph Theory}

Graphs are combinatorial structures of configurations of nodes and connections \cite{Gross2018GraphApplications}. Graph theory is a prominent field within discrete mathematics, studying the properties of graph structures. The two following definitions are a predominant graph theory textbook, updated in 2019, by Gross, Yellen, and Anderson titled \textit{Graph Theory and Its Applications}.

\subsection{Graph Formalism}

\begin{definition}
A \textbf{graph} $G = (V, E)$ is a mathematical structure consisting of two finite sets $V$ and $E$. The elements of $V$ are called vertices (or nodes), and the elements of $E$ are called edges. Each edge has a set of one or two vertices associated to it, which are called its endpoints.

\end{definition}

An edge is said to join its endpoints. A vertex joined by an edge to a vertex $v$ is said to be a neighbour of $v$.

\begin{definition}
The (open) \textbf{neighbourhood} of a vertex $v$ in a graph $G$, denoted $N(v)$, is the set of all the neighbours of $v$. The closed neighbourhood of $v$ is given by $N[v] = N(v) \cup \{v\}$.
\end{definition}

When $G$ is not the only graph under consideration, the notations $V_G$ and $E_G$ (or $V(G)$ and $E(G)$) are used for the vertex- and edge-sets of $G$, and the notations $N_G(v)$ and $N_G[v]$ are used for the neighbourhoods of $v$.

\subsection{Graph Similarity Metrics}

Quantifying the similarity between two graphs is a complex problem. This Graph Similarity Metrics section discusses a common metric for comparing graphical structure followed by two metrics well-established means of comparing sets of points in a metric space which can be applied to graphs.

\subsubsection{Graph Edit Distance}

The Graph Edit Distance between two graphs $G_1$ and $G_2$ is defined as the minimum cost sequence of edit operations required to transform $G_1$ into $G_2$ \cite{Riesen2009ApproximateMatching}. The edit operations can involve insertion, deletion, or substitution of vertices or edges, each with an associated cost. The goal is to find the sequence of operations with the minimum total cost \cite{Riesen2009ApproximateMatching}. 

\begin{definition}
    The \textbf{Graph Edit Distance} is given by 

$$
GED(G_1, G_2) = \min_{\phi: V_1 \cup E_1 \rightarrow V_2 \cup E_2} \sum_{x \in V_1 \cup E_1} c(x, \phi(x))
$$

where $\phi: V_1 \cup E_1 \rightarrow V_2 \cup E_2$ is a partial function that maps vertices and edges of $G_1$ to vertices and edges of $G_2$, and $c(x, \phi(x))$ is the cost of substituting, deleting, or inserting $x$ depending on $\phi(x)$.
\end{definition}

The GED problem is NP-hard and correspondingly computationally difficult to solve exactly \cite{Incudini2022ComputingDevices}.

\subsubsection{Wasserstein Distance}

The Wasserstein distance between two probability measures on a metric space, also known as the Earth Mover's distance, is a measure of the optimal cost of transporting mass to transform one measure into the other \cite{Villani2007OptimalNew}. In the context of graph theory, this can be extended to the space of graph structures by considering the graphs as distributions of features \cite{Peyre2019ComputationalTransport}.

\begin{definition}
    The $p$-th order \textbf{Wasserstein distance} between two probability measures $\mu$ and $\nu$ on a metric space $(M, d)$ is defined as:

$$
W_p(\mu, \nu) = \left( \inf_{\gamma \in \Gamma(\mu, \nu)} \int_{M \times M} d(x, y)^p \: d\gamma(x, y) \right)^{1/p}
$$

where
    \begin{itemize}[noitemsep]
        \item $W_p(\mu, \nu)$ is the $p$-th order Wasserstein distance.
        \item $d(x, y)$ is the given metric on the space $M$.
        \item $\Gamma(\mu, \nu)$ is the set of all joint distributions $\gamma(x, y)$ whose marginals are $\mu$ and $\nu$ respectively.
        \item The $\inf$ is the infimum over all possible such $\gamma$.
    \end{itemize}
\end{definition}

Calculating the Wasserstein distance can be computationally expensive, especially for high-dimensional data \cite{Peyre2019ComputationalTransport}.

\subsubsection{Hausdorff and Relative Hausdorff Distances}

The Hausdorff distance is a measure commonly used in the field of computer vision and pattern recognition. It allows for a many-to-many correspondence between two sets, often making it more flexible for comparison tasks than measures requiring a one-to-one mapping \cite{Huttenlocher1993ComparingDistance}.

\begin{definition}
    Given two point sets $A = {a_1, a_2, ..., a_n}$ and $B = {b_1, b_2, ..., b_m}$, the one-sided \textbf{Hausdorff distance} from $A$ to $B$ is given by

$$
\tilde{\delta}_{H} (A, B) = \max_{a \in A} \min_{b \in B} |a - b|
$$

\end{definition}

This measure can be computed on Voronoi diagrams in $O((n + m) \log(n + m))$ time \cite{Edelsbrunner2002TopologicalSimplification}.

The Relative Hausdorff distance is a variant of the Hausdorff distance that is especially suitable for comparing graphs. It was developed to measure the closeness of degree distributions at all scales, which makes it an effective tool for comparing the heavy-tailed degree distributions frequently observed in real-world graphs \cite{Aksoy2019RelativeAnalysis}. 

\begin{definition}
    The \textbf{Relative Hausdorff distance} between two graphs, $\Upsilon$ and $\Lambda$, is a measure of closeness between their complementary cumulative degree histograms. The metric is given by \\
$$
\Tilde{\delta}_{RH} (\Upsilon, \Lambda) = \max\{ \Tilde{\delta}_{RH}\rightarrow(\Upsilon, \Lambda), \Tilde{\delta}_{RH}\rightarrow(\Lambda, \Upsilon) \}
$$

where $\Tilde{\delta}_{RH}\rightarrow(\Upsilon, \Lambda)$ is the minimum $\epsilon$ such that $ \forall \; d \in \{1, \dots,\Delta(\Upsilon)\}, \;  \exists \; d' \in \{1, \dots,\Delta(\Lambda)+1\} $ such that $|d-d'| \leq \epsilon d$ and $|\Upsilon(d)-\Lambda(d')| \leq \epsilon \Upsilon(d)$. 
\end{definition}

The Relative Hausdorff distance is less computationally intensive than the Hausdorff distance, especially for large graphs or graphs embedded in high-dimensional spaces, and can capture the degree distribution closeness at all scales \cite{Aksoy2019RelativeAnalysis}.

\subsubsection{Dual Graph Formalism}

A dual object in mathematics is a transformation that turns one kind of mathematical object into another kind, preserving the relationships and characteristics of the original.

\begin{definition}
    The \textbf{dual graph} $G^*$ of the planar graph $G$ is defined as the union of a vertex set, denoted as $V(G^*)$, and an edge set, denoted as $E(G^*)$, where:
\begin{itemize}[noitemsep]
    \item The vertex set of $G^*$ is in one-to-one correspondence with the face set of $G$, $F(G)$.
    \item The edge set of $G^*$ is in one-to-one correspondence with the edge set of $G$, $E(G)$. An edge in $G^*$ connects two vertices (corresponding to faces in $G$) if and only if the corresponding faces share a common edge in $G$.
\end{itemize}
\end{definition}

This translation expresses the essence of the original in a different form.
In graph theory, the concept of a dual graph is applied to planar graphs, graphs that can be embedded in the plane without edge crossings \cite{Kay1977GraphApplications}.

\subsection{Rotation Systems}

\begin{figure}[!htb]
\centering
\includegraphics[width=0.95\textwidth]{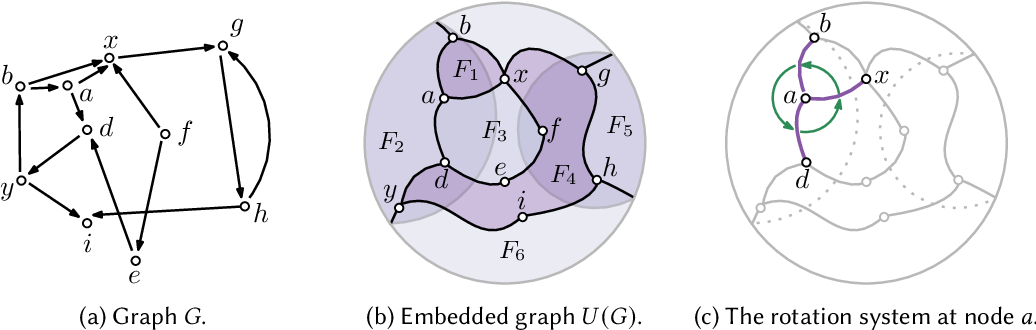}
\caption{(a) the drawing of a graph $G$ with edge crossings. (b) representation of the graph $G$ embedded in the sphere is shown in. The graph embedded $U(G)$ serves as the symmetrisation of the graph $G$ where an edge $e$ in $G$ from $x$ to $y$ induces an edge in $U(G)$ from $x$ to $y$ and an edge from $y$ to $x$. The corresponding faces of the graph embedding shaded are named $F_i$ for $i$ from 1 to 6. (c) the incident edges at the node $a$ in $U(G)$ in purple, the rotation system at $a$ in green. }
\label{fig:rotation}
\end{figure}

In combinatorial mathematics, rotation systems, also called combinatorial embeddings or combinatorial maps, encode embeddings of graphs onto orientable surfaces by describing the circular ordering of a graph's edges around each vertex \cite{Wikipedia2023RotationSystem}. Rotations systems are also known as combinatorial embeddings or combinatorial maps and their formal definition is given by Definition 12 \cite{Wikipedia2023RotationSystem} followed by a visualisation in Figure \ref{fig:rotation} \cite{Prieto-Cubides2021OnTheory}.

\begin{definition}
    A \textbf{rotation system} is defined as a pair $(\sigma, \theta)$ where $\sigma$ and $\theta$ are permutations acting on the same ground set $\beta$, $\theta$ is a fixed-point-free involution, and the group $<\sigma, \theta>$ generated by $\sigma$ and $\theta$ acts transitively on $\beta$.
\end{definition}

Combinatorial maps are dimension-independent and rely on a single element along with a simple set of relations. All the information about the cells and their incidence and adjacency relations is contained within this simple model. All neighbourhood queries are resolved in optimal time, linear in the number of traversed cells, without having to maintain any additional information \cite{CGoGN2023CombinatorialMaps}.


\section{Computational Geometry}

Computational Geometry is a field of study dedicated to the resolution of geometric problems.

\subsection{Voronoi Diagram}

A foundational structure of interest in computational geometry are Voronoi diagrams.
A Voronoi diagram is a geometric partitioning of a space into regions, where each region is associated with a unique generating point, commonly called a "site", such that every location within a region is closer to its corresponding generating point than to any other generating point \cite{Aurenhammer1991VoronoiStructure}. In essence, it is a tessellation that optimally assigns each point in the space to its nearest site, providing a natural representation of spatial proximity and neighbour relationships.

\begin{definition}
Let $S=\{s_1, s_2, \dots, s_m\}$ is a set of $s$ points in our $n$-dimensional space. \textbf{Voronoi cells} $V_i$ is a cell that contains all the points closer to $s_i \in S$ than to any other point.
The Voronoi cell of a site $s_i \in S$, denoted as $V(s_i)$, given by

$$
    V(s)_i=\{x: \forall j \neq i, d(x, s_i) \leq d(x, s_j)\}, \quad \text{with} \quad i,j \in \{1, 2, \dots, c\}
$$
using the concept of distance function $d(x, y)$, which gives the distance between points $x$ and $y$ in the plane.

The \textbf{Voronoi Diagram} is the union of all Voronoi cells for points in $S$:

\begin{equation}
    \mathcal{V}(S) = \bigcup_{s \in S} \; \{ V(s) \}
\end{equation}

\end{definition}

The distance function $d(x, y)$ is used to calculate the distance between points. Predominately, the Euclidean distance (l$_2$ distance) is used:

\begin{equation}
    d(x, y) = \lVert x - y \rVert_2 = \sqrt{\sum_{k=1}^n (x_k - y_k)^2}
\end{equation}

However, other distance metrics, such as Manhattan distance (l$_1$ distance), are used as well, and others required for higher dimensions.

\subsubsection{Geometric Explanation}

\begin{figure}[htb]
\centering
\includegraphics[width=1\textwidth]{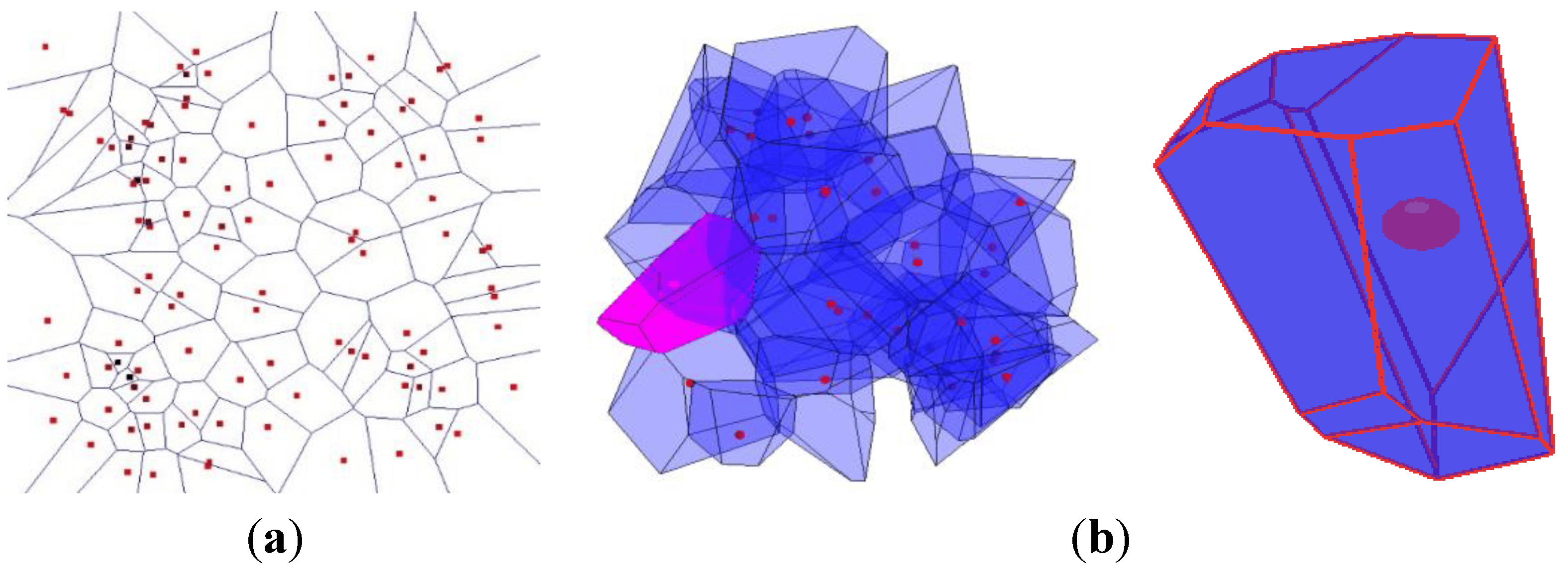}
\caption{Two-dimensional (a) and three-dimensional (b) images of Voronoi diagrams from a 2015 article by Ying, Xu, Li, and Mao titled \textit{Point Cluster Analysis Using a 3D Voronoi Diagram with Applications in Point Cloud Segmentation}.}
\label{fig:2d3dvor}
\end{figure}

Consider a set of sites in the plane. For each seed, the Voronoi cell is the region of all points that are closer to that seed than any other sites \cite{Aurenhammer1991VoronoiStructure}. The boundaries of Voronoi cells are equidistant from neighbouring sites, neighbours as given by Definition 7, and the vertices of the diagram are equidistant from three or more seeds which is visualised in Figure \ref{fig:2d3dvor} below \cite{Ying2015PointSegmentation}.

\subsubsection{Properties of Voronoi Diagrams}

The Voronoi diagram has several notable properties:

\begin{enumerate}
    \item Coverage: The union of all Voronoi cells covers the entire metric space, $\bigcup_{s_i \in S} V(s_i) = M$.
    \item Partition: The Voronoi cells are disjoint, $V(s_i) \cap V(s_j) = \emptyset$ for any $s_i, s_j \in S$ with $i \neq j$.
    \item Adjacency: Two sites $s_i, s_j \in S$ are said to be adjacent if their corresponding Voronoi sites $V(s_i)$ and $V(s_j)$ share a non-empty common border. The adjacency relationship between seeds induces a graph structure on the set of sites, known as the Voronoi graph of $S$.
    \item Convexity: If the metric space is the Euclidean space and the distance function is the Euclidean distance, then each Voronoi cell is a convex set \cite{Aurenhammer1991VoronoiStructure} \cite{Dobrin2005ADiagrams}.
\end{enumerate}

\subsubsection{Delaunay Triangulation}

\begin{figure}[htb]
\centering
\includegraphics[width=1\textwidth]{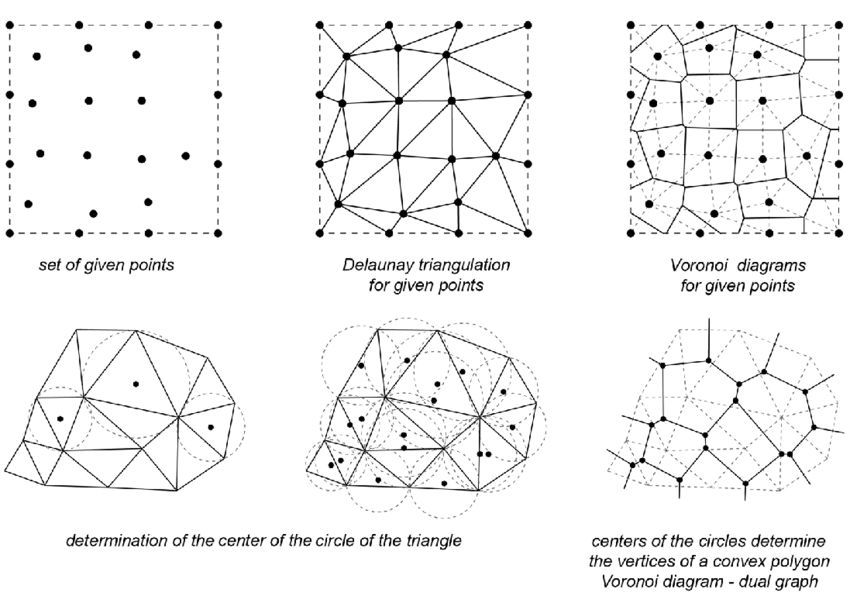}
\caption{Visual of Delaunay Triangulation and Voronoi Diagram created from a set of points from a 2016 article by Rokicki and Gawell titled \textit{Voronoi diagrams – rod structure research models in architectural and structural optimisation}.}
\label{fig:delauney}
\end{figure}

Delaunay triangulation gives the dual graph to its associated Voronoi diagram. Delaunay triangulation is a technique used in computational geometry to create a triangulation of a set of points in the plane such that each edge connects two points with adjacent Voronoi cells (URL) \cite{Dobrin2005ADiagrams}). This can be seen in \ref{fig:delauney} below \cite{Rokicki2016VoronoiOptimization}.

Notice that given a set of points $P = \{p_1, p_2, \ldots, p_n\}$ in the plane, a triangulation is a division of the convex hull of the points into non-overlapping triangles, such that each point in $P$ is a vertex of one or more triangles \cite{DeBerg2008ComputationalApplications}. This triangulation has the property that the circumcircle of any triangle formed by an edge in the graph contains no other seeds in its interior \cite{Aurenhammer2000Voronoi2-2.}. This triangulation maximises the minimum angle of all the triangles in the triangulation, avoiding triangles with very sharp angles \cite{Lee1980TwoTriangulation}. This property can be seen in \ref{fig:delauney} and makes Delaunay triangulation useful in many applications such as mesh generation and spatial interpolation. Several algorithms exist for their construction, including the incremental algorithm, divide-and-conquer algorithm, and the sweep-line algorithm \cite{DeBerg2008ComputationalApplications}.

\subsection{Stability and Quantifying Change}

The concept of stability or sensitivity in computational geometry involves measuring how small perturbations to the input (the set of points) affect the output (the Voronoi diagram, Delaunay triangulation, or their dual graph). Small perturbations to the set of points can lead to changes in the structure of the Delaunay triangulation or Voronoi diagram, with the nature and extent of these changes being dependent on the specific configuration of the points and the nature of the perturbation \cite{Edelsbrunner2002TopologicalSimplification}. 

\subsubsection{Relationship between Changes in the Generating Sites and Changes in the Resulting Diagrams}

\begin{figure}[!htb]
\centering
\includegraphics[width=1\textwidth]{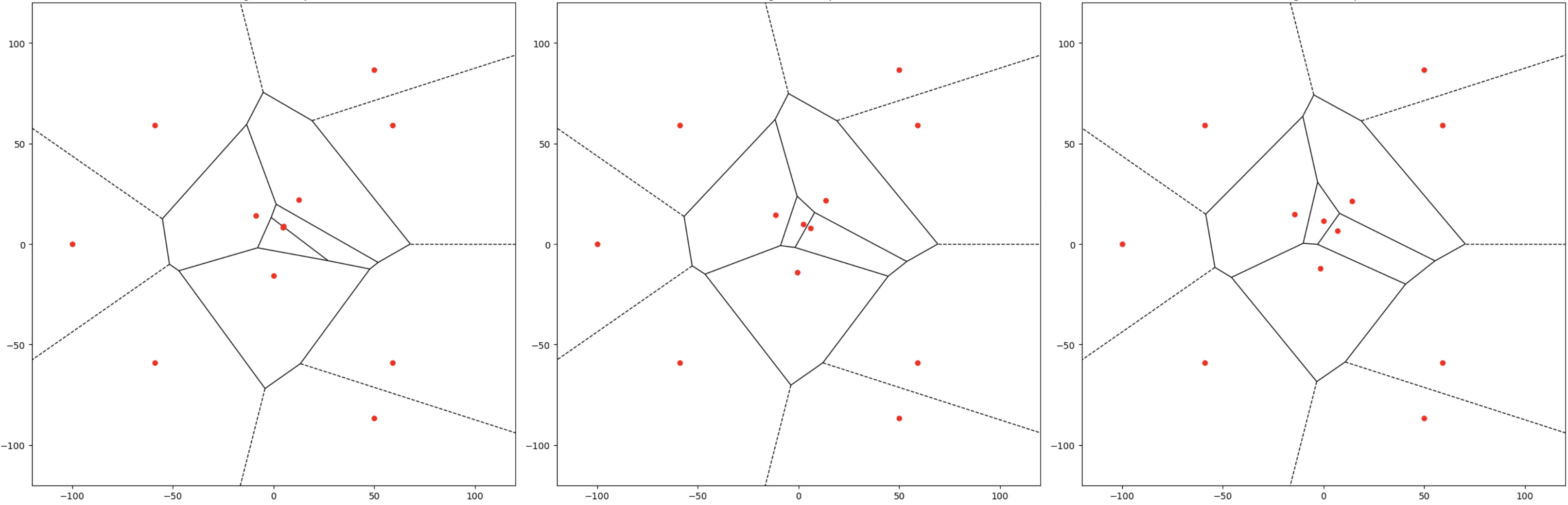}
\small{(a) \quad \quad \quad \quad \quad \quad \quad \quad \quad \quad \quad \quad \quad \quad (b) \quad \quad \quad \quad \quad \quad \quad \quad \quad \quad \quad \quad \quad \quad (c)}
\caption{Figure (a) is of the base Voronoi diagram in black resulting from the sites shown in red. Figure (b) shows red sites, the five central most have been perturbed, using Voronoi site generation procedure outlined in the beginning of this section, by $\alpha = 0.05$, and showing the resulting Voronoi diagram that is differing in structure to that of (a). Figure (c) is figure (b) with an additional $0.05$ perturbation making $\alpha = 0.1$ yet only resulting in warping of the resulting diagram but no connection-changing structural change to that of (b).}
\label{fig:similarity}
\end{figure}

The relationship between the changes in sites that generate diagrams and the resulting Voronoi diagrams, is strictly monotonic. Dependent on the location of the sites, small perturbations in the sites positions can lead to significant changes in the Delaunay triangulation, such as edge flips, and consequently, in the Voronoi diagram \cite{Guibas1983PRIMITIVESDIAGRAMS.}. Perturbations may also warp the corresponding Voronoi diagram while not effecting the graphical structure of the diagram at all if there remains any site with the relative position of the previous nearest neighbours as before. This relationship is depicted in Figure \ref{fig:similarity}

 As the impact of a perturbation depends on the specific configuration of the points and the nature of the perturbation, quantifying this effect in a general and measurable way still presents some challenges.

\chapter{Related Work}

\section{Catastrophic Forgetting}

Most work in understanding catastrophic forgetting has been in the domains of supervised learning and neural networks.
Of the research on catastrophic forgetting, studies have concentrated on developing methods for mitigating forgetting with techniques to regulate the change of weights in the network \cite{Kirkpatrick2017OvercomingNetworks, Zenke2017ContinualIntelligence}, the experience replay \cite{Fazeli-Asl2023ADiscrepancy, Fedus2020OnGames, Hafez2023Map-basedLearning, Mnih2015Human-levelLearning, Mnih2013PlayingLearning, Parisi2018ContinualReview, Schaul2015PrioritizedReplay, vandeVen2018GenerativeLearning}, and the local optimisation of neural networks \cite{Fedus2020OnGames, Ghiassian2020ImprovingNetworks, Liu2018TheLearning} as well as meta learning \cite{Baik2019LearningMeta-Learning, Spigler2019Meta-learntNetworks}.

However, there has been limited work on how forgetting is influenced by the similarity, or dissimilarity, of the new task from the initial, which is the focus of this paper.

\section{Effect of Task Similarity on Catastrophic Forgetting}

The 2013 paper "An Empirical Investigation of Catastrophic Forgetting in Gradient-Based Neural Networks" analysed the impact of similar and dissimilar tasks on catastrophic forgetting in gradient-based neural networks. This investigation was carried out across various gradient-based training algorithms and using different activation functions. Although a difference in forgetting was identified based on the difference of the new task, this effect was inconsistent across the different activation functions \cite{Goodfellow2013MaxoutNetworks}.

A notable study in 2020, "Anatomy of Catastrophic Forgetting: Hidden Representations and Task Semantics", observed maximal forgetting in neural network models occurs for task sequences with intermediate similarity \cite{Ramasesh2020AnatomySemantics}.

The 2021 paper, "Continual Learning in the Teacher-Student Setup: Impact of Task Similarity" primarily aimed to investigate the effect of task similarity on forgetting, providing the most recent and direct work on this relationship. This research examined the problem of catastrophic forgetting in the context of deep learning networks and found that the greatest forgetting occurs at an intermediate task similarity, when tasks depend on similar features \cite{Lee2021ContinualSimilarity}.

\section{Task Similarity Metrics in Reinforcement Learning}

In the pursuit of a more comprehensive and accurate metric for task similarity in RL, a recent advance in the performance-based metrics category learns an embedding for sequential tasks as a similarity measure. Proposed this year in "Learning Embeddings for Sequential Tasks Using Population of Agents", the embeddings are based on an information-theoretic framework. The information-theoretic criterion is directed by the concept that two tasks can be considered similar if observing an agent's performance on one task can reduce our uncertainty about its performance on the other \cite{Mahajan2023LearningAgents}.

\chapter{Methodology}


To explicate the relationship between task similarity and forgetting, a way to generate related tasks, a metric for measuring the difference between tasks, and a method for quantifying the degree of forgetting after sequential training are needed. A tabular, Q-learning method with the objective of minimising the steps taken to reach a goal is used; details of which follow in the 4.3 Agent section. 

To enable a clear quantification of task similarity, the tasks are based on graphs. The use of graphs avoids some of the subjectivity, discussed in Section 2.3, in similarity metrics. Voronoi diagrams, defined in Section 2.5.1, offer a means for generating similar graphs by manipulating the original sites locations. By optimising for the number of steps taken, the agent interprets a difference in tasks by only the graph's topological structure, irrespective of edge length or spatial distortion. To measure task similarity, a metric for structural difference between graphs, which is computationally feasible, is established. Details of the generation and measurement of similar tasks are in the 4.2 Task section. 

The extent of forgetting is quantified by the difference in the performance from training on an initial task, and the performance after training on a new task. Each RL  agent is trained on a base graph with the step-minimising objective and a specified goal state. The performance is measured by the number of steps taken, given a specified starting state. Next, the agent is trained on a new graph with the same specified goal state (details of enabling consistent goal state are in the 4.2 Task section). Then, the performance on the base graph, with the same starting state, is measured. The difference between the two performances is the measure \textit{forgotten}. The details of capturing these measures follow in the 4.4 Evaluation section.

The experimental trials, the parameters tested, and the measurements tracked are expressed in the 4.4.3 Experiment section.

\section{Tasks}
This section presents the systematic approach employed for the creation of tasks, with a focus on ensuring the tasks remain related. This is followed by the means used to evaluate and quantify the degree of task similarity.

\subsection{Generating Tasks}

The procedure for creating similar tasks is accomplished by generating graphs from Voronoi diagrams with sites manipulated by a controlled perturbation of base sites. 

\subsubsection{Voronoi Diagram Site Generation}

Voronoi diagrams serve as a geometric tool to divide a given space into several regions based on a specific set of sites, as defined in Section 2.5.1. The set of sites $S$ is comprised of two disjoint subsets: \textit{external} points denoted as $N \in \mathbb{R}^2$, and \textit{internal} points denoted as $M \in \mathbb{R}^2$. 

To provide a consistent outer perimeter for the graphs of the resulting Voronoi diagrams, the set of external sites will remain the same. These external points are evenly distributed along the circumferences of two concentric circles centered on the origin. On the larger of the two circles, given by radius $R$, the \textit{outer} external sites are allocated. On the smaller of the two circles, with radius $r$, the \textit{inner} external sites are allocated. 

Enabling the generation of many base diagrams, the internal sites are sampled from a normal distribution that centers at the center of the external sites. The multivariate Gaussian distribution of the internal sites is given in equation \ref{equation:Gaussian}.

\begin{equation}
M \sim \mathcal{N}\left((0,0), \left(\frac{R}{4}\right)^2 \mathbb{I}\right)
\label{equation:Gaussian}
\end{equation}

To ensure the resulting Voronoi diagrams from different internal sites change while remaining related, two random sets of inner sites are sampled, $x_0 \in M$ and $x_1 \in M$, and interpolating between the sets using equation \ref{equation:perturbation}:

\begin{equation}
x_\alpha = \cos\left(\frac{\pi}{2} \alpha\right) x_0 + \sin\left(\frac{\pi}{2} \alpha\right) x_1
\label{equation:perturbation}
\end{equation}

where $\alpha \in [0,1]$. Notably, because the interpolation is a linear combination of Gaussian-distributed points in such a way that preserves the variance, it is ensured that the the interpolated tasks retain the same distributional properties of the original distributions. The parameter $\alpha$ can be interpreted as the fraction of the variance, providing a mechanism to control the difference between sites. When $\alpha$ is 0, the task is entirely defined by $x_0$; as $\alpha$ increases to 1, the influence of $x_1$ grows while the influence of $x_0$ diminishes correspondingly.

Together, the outer external, inner external, and internal points serve as the sites for the Voronoi diagram.

\subsubsection{Voronoi Diagrams Generation}

To create these diagrams, the scipy.spatial.Voronoi functions from the SciPy library in Python was used. This function relies on the Qhull library, a robust computational framework for tackling spatial problems such as Voronoi diagrams, Delaunay triangulations, and convex hulls.

\begin{figure}[!hbt]
\centering
\includegraphics[width=1\textwidth]{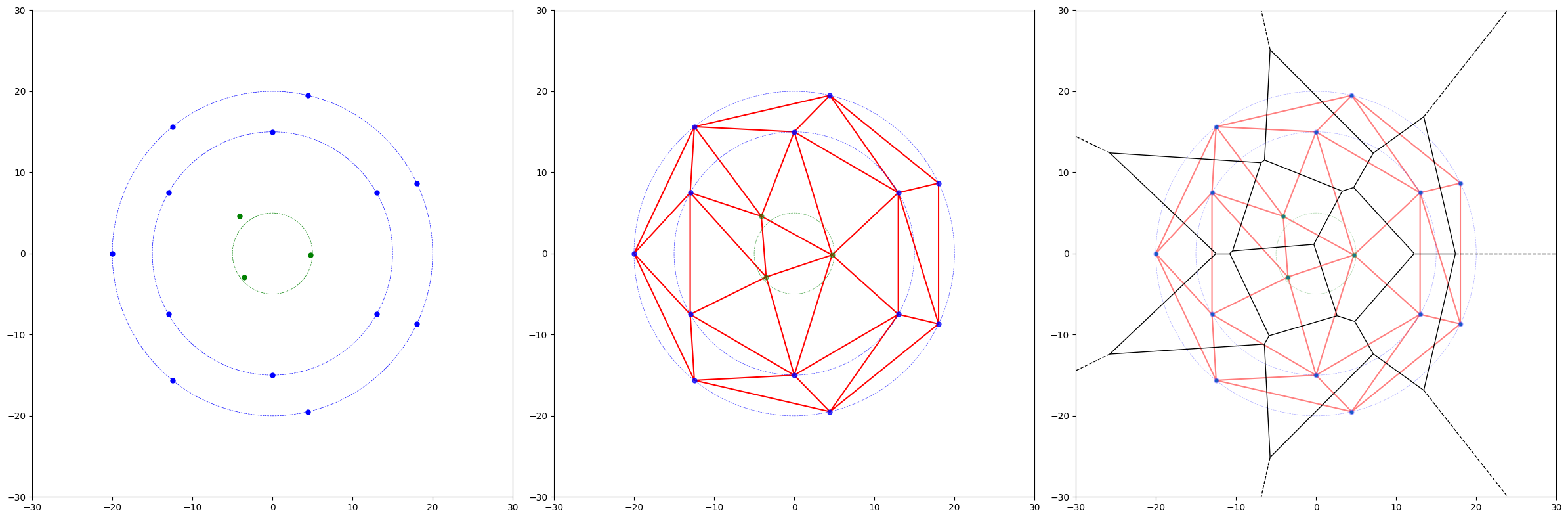}
\scriptsize{(a) External and Internal Sites $S$ \quad \quad (b) Delaunay Triangulation $D(S)$ \quad \quad (c) Resulting Voronoi Diagram $\mathcal{V}(S)$}
\caption{(a) shows $7$ \textit{outer} external sites, $6$ \textit{inner} external sites, and $3$ internal sites. The external sites, $N$ and the corresponding circles they are evenly placed upon are shown in blue. The internal points $M$ and a circle marking one standard deviation from the center of the Gaussian from which the points are generated are shown in green. (b) adds to (a) the Delaunay triangulation of the sites $S = N \cup M$. (c) adds to (b) the boundaries and intersections of the Voronoi polygons, giving the resulting Voronoi diagram for the sites $S$.}
\label{fig:create_vor}
\end{figure}

\newpage
The specific process Qhull employed to generate a Voronoi diagrams, using Delaunay triangulation, proceeds through these steps:

\begin{enumerate}
    \item \textit{Delaunay Triangulation:} 
    The Delaunay triangulation $D(S)$ of the input sites $S$ is computed. This is visualised in Figure \ref{fig:create_vor}b, with a red Delaunay triangulation of the blue and green sites.
    \item \textit{Circumcenter Calculation:}
    The circumcenter $c_{ijk}$ for each triangle $\triangle_{ijk} \in D(S)$ is calculated. These circumcenters serve as the vertices $V = {v_{ijk}}$ of the Voronoi diagram.
    \item \textit{Boundaries:}
    The boundaries $B = {b_{ij}}$ of the Voronoi regions are formed by the set of points equidistant from two sites $s_i, s_j \in S$. The process employs Euclidean distance to create a patchwork of polygons that blanket the entire space. Each intersection of two Voronoi cell boundaries $b_{ij}, b_{jk} \in B$ creates an edge $e_{ijk} \in E$ which become the edges of the Voronoi diagram.
    \item \textit{Resulting Diagram: }
    This process results in the Voronoi diagram $\mathcal{V}(S) = {R_i, E, V}$, where $R_i$ are the regions, $E$ are the edges, and $V$ are the vertices. Figure \ref{fig:create_vor}c shows the resulting Voronoi diagram in black, from the red Delaunay triangulation of the blue and green sites.
\end{enumerate}

\subsubsection{Consistent Vertex Label Assignment across Corresponding Voronoi Diagrams}

Building upon the duality of Voronoi diagrams and Delaunay triangulations, a strategy is created that assigns labels to the vertices consistently across different Voronoi diagrams and their corresponding Delaunay triangulations. Each vertex is one-hot encoded, signifying the binary representation of the Delaunay triangle associated with the vertex. To maintain a consistent mapping between sites, Voronoi vertices, and Delaunay triangles across varying diagrams, a method rooted in computational geometry and graph theory was employed. To efficiently organise points in the multi-dimensional space, a k-d tree data structure is constructed from the original set of sites. This structure partitions the space into various regions, enabling rapid nearest neighbour queries.

For each subsequent Voronoi diagram with site points $S_{\alpha} = {\mathbf{s}_{\alpha,1}, \mathbf{s}_{\alpha, 2}, ..., \mathbf{s}_{\alpha, |N|+|M|}}$, where $\alpha = 0$ gives the initial Voronoi diagram, a k-d tree is utilised to find the closest original site $\mathbf{s}_{0,j}$ for each site $\mathbf{s}_{\alpha, i}$. This step creates a mapping from the perturbed sites back to the original.

For every vertex $\mathbf{v}_{\alpha, i}$ in the set of vertices $V = {\mathbf{v}_{\alpha, 1}, \mathbf{v}_{\alpha, 2}, ..., \mathbf{v}_{\alpha, \eta}}$, the Euclidean distances to all sites in $S$ are computed and the three nearest sites identified. Formed from these three nearest sites is the Delaunay triangle $\triangle_{ijk}$ corresponding to the Voronoi vertex $\mathbf{v}_{\alpha, i}$. 

This association between Voronoi vertices, Delaunay triangles, and original sites is represented by a binary-encoded matrix $B \in {0, 1}^{m \times p}$. Each entry $b_{ij}$ in the matrix is assigned a value as given in Equation \ref{equation:one-hot}.

\begin{equation}
b_{ij} =
\begin{cases}
1, & \text{if } \triangle_{ijk} \text{ corresponds to } v_{\alpha, i} \text{ and } \exists \; k \text{ s.t. } I_{k} = j \\
0, & \text{otherwise}
\end{cases}
\label{equation:one-hot}
\end{equation}

\subsubsection{Extracting a Network Graph from the Voronoi Diagram}

A graph, denoted as $G=(\mathcal{N},\mathcal{E})$, where $\mathcal{N}$ is the set of nodes and $\mathcal{E}$ is the set of edges, is formed based on the Voronoi diagrams. The Voronoi vertices $V$ in the Voronoi diagram $\mathcal{V}(S)$ corresponds to a nodes $N$ in $\mathcal{N}$ of the graph $G$. The node $n \in \mathcal{N}$ is assigned the vector of coordinates $(x,y)_{n}$ of the corresponding Voronoi vertex $v \in V$. The labelling of the node $n$ is determined by the one-hot encoding labelling method given previously in Section 4.2.3. For every finite Voronoi edge $e^{finite}$ in $E(S)$, there is a corresponding edge $\varepsilon \in \mathcal{E}$ of the graph $G$. The infinite edges of the Voronoi diagram are excluded from the graph.

The resulting graph $G$ has a structure that resembles the connected portion of the Voronoi diagram and incorporates the consistent labelling between \textit{similar} graph (i.e. the graphs resulting from perturbation).

\subsection{Measuring Similarity}

The procedure for measuring the similarity between tasks is accomplished by a (feasible) metric that indirectly captures the structural differences between two graphs.

The difference measure of interest is that of the topological structure of two graphs. To reiterate, purely the connections of the graph is of interest due to the agent's step-based optimisation. A change in the length of the edges traversed or a change to the angles at which the agent is moving is of no consequence in the interpretation of the task by this agent if the change does not alter the connections of the graph.

Recall from 2.5.2, the extent of structural changes does not necessarily correlate with the magnitude perturbing the sites because the topological structure depends not solely on each site's positions but also their relative position to each other. It is for this reason the similarity between two tasks is computed using a metric focused on the topological structures between graphs.

Although calculating the Graph Edit Distance, defined in Section 2.4.2, provides a direct measure of the structural difference between two graphs, it is computationally expensive and impractical for larger graphs due to its NP-hard nature. This makes it, unfortunately, beyond the capability of computation in this study. The Wasserstein distance and the Hausdorff distance offer more computationally feasible alternatives. 

The Wasserstein metric, defined in Definition 9, captures the overall difference in the spatial distribution of the cells. Let the Voronoi diagram associated with the graph $G$ be $\mathcal{V}^G$, and the area of the Voronoi cell corresponding to a vertex $v \in V$ be $a(v)$. Let $\Gamma(G_1, G_2)$ denotes the set of all joint distributions over vertex pairs $(v_1, v_2)$ for $v_1 \in V^G$, whose marginals correspond to the distributions of areas in $\mathcal{V}^G$. The second order Wasserstein distance between the Voronoi diagrams that gave the two graphs $G_1$ and $G_2$, calculated over the areas of Voronoi cells is given in Equation \ref{equation:wass_graph}.

\begin{equation}
    W_2(G_1,G_2) = \left( \inf_{\gamma \in \Gamma(G_1, G_2)} \int_{\mathcal{N}_1 \times \mathcal{N}_2} d(a(v_1), a(v_2))^2 \: d\gamma(v_1, v_2) \right)^{1/2}
    \label{equation:wass_graph}
\end{equation}

The computation of the Hausdorff distance defined in Definition 10, between the graphs $G_1(\mathcal{N}_1, \mathcal{E}_1)$ and $G_2(\mathcal{N}_2, \mathcal{E}_2)$, is given in Equation \ref{equation:haus_graph}.

\begin{equation}
    \delta_H(G_1, G_2) = \max\left(\max_{n_1 \in \mathcal{N}_1} \min_{n_2 \in \mathcal{N}_2} ||n_1 - n_2||, \max_{n_2 \in \mathcal{N}_2} \min_{n_1 \in \mathcal{N}_1} ||n_2 - n_1||\right)
    \label{equation:haus_graph}
\end{equation}

These metrics are used in combination to provide a relatively comprehensive measure of the differences between the topological graph structures. To combine these metrics, the average of the normalised Wasserstein and Hausdorff distances is computed. The combined metric, $\varDelta(G_1, G_2) = \big | (W_2(G_1, G_2) + \delta_H(G_1, G_2)) / 2 \big |$, provides a measure that indirectly captures the structural differences between two graphs. This metric is used for measuring task similarity.

\section{Agent}

The agent operates on a closed, undirected graph derived from a Voronoi diagram. Each node of the graph represents a discrete state, and the edges signify actions that the agent can execute. The objective of the agent is to traverse the graph in such a way as to reach the goal state in the fewest possible steps. For interpretability, the agent uses tabular Q-learning, interacting with its environment to incrementally learn the values of actions for each state. The Q-learning algorithm operates in an iterative manner, balancing exploration of new actions and exploitation of known information. This balance is achieved using an $\epsilon$-greedy strategy with a decaying $\epsilon$-value. The agent updates a Q-table, from which it derives a policy that is used to make decisions to maximise the cumulative reward. Note that the action-selection process is based on the graph, not on the Q-table which is not graph-specific. This flexibility allows the agent to handle training on new task (i.e. graphs) to which the Q-table may not have been initialised.

\subsection{Initialisation}

States $S \in \mathcal{S}$ are identified according to the labels of the nodes of the Voronoi graphs. Each state has a set of available actions, $A \in \mathcal{A}$, which correspond to transitions to neighbouring states along an edge. The state transitions, $P \in \mathcal{P}$, are deterministic.

To allow the agent consistency in its interact with difference graphs, the actions are constructed with a rotation system. The rotation system provides a means of abstracting the graph's geometry to shift reliance on the specific structure to the combinatorial properties of the graph. To create this system, a sorted list of states and its neighbours is assembled into a state-neighbour dictionary. This state-neighbour dictionary is arranged such that the neighbouring states associated with each state are sorted in clockwise order relative to due north of the state. This results in a consistent and ordered labelling of neighbouring states for each state in the dictionary that preserves a notation of relative position of its neighbouring states. Index $0$ of the ordered neighbouring states corresponds to the upper-right most neighbouring state, the neighbour at index $1$ to the neighbour in the clockwise direction, and the neighbour at index $2$ to a third neighbouring state at a greater 'time' on the theoretical clock face. 

\begin{figure}[!htb]
    \centering
    \includegraphics[width=0.49\textwidth]{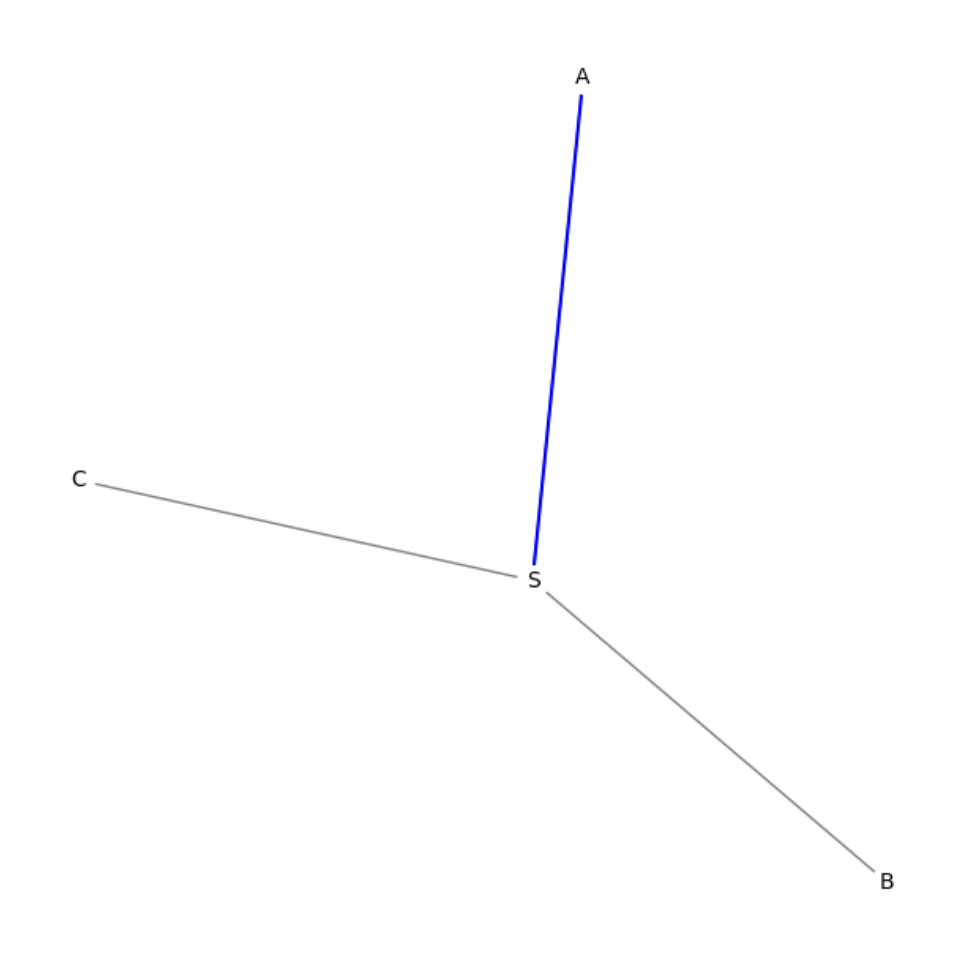}
    \includegraphics[width=0.49\textwidth]{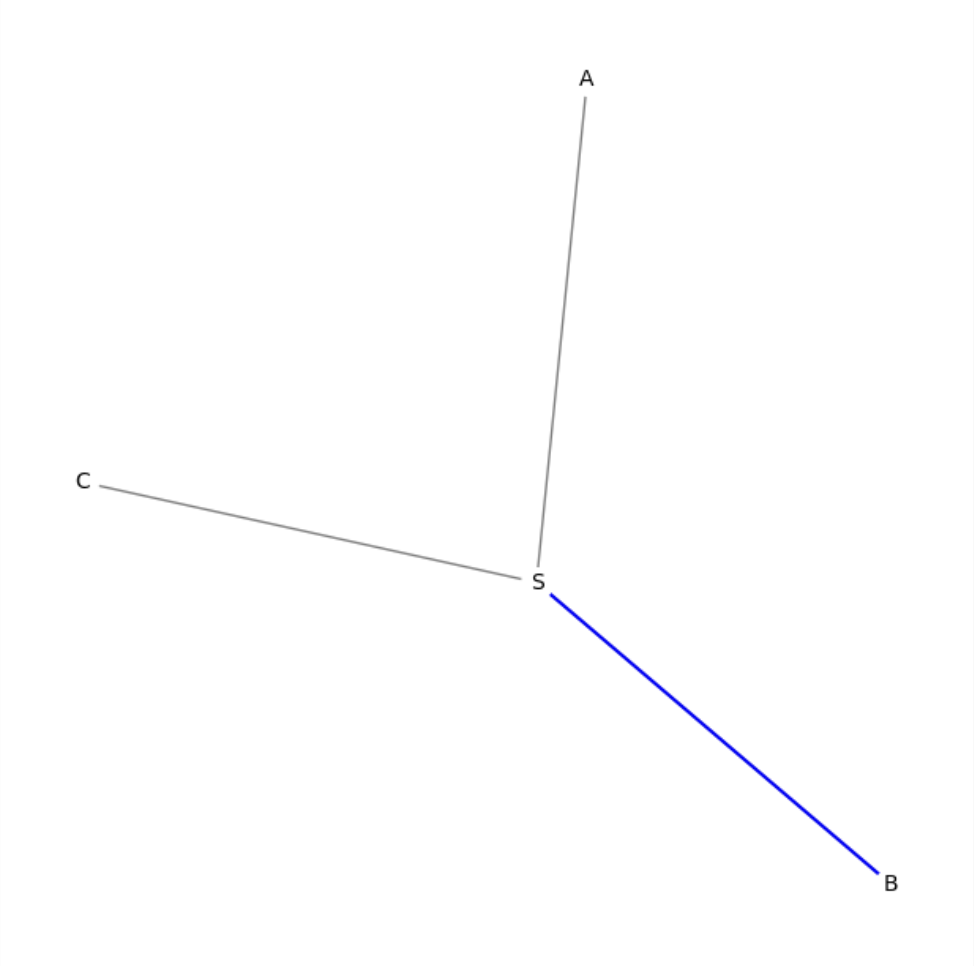}
    \includegraphics[width=0.49\textwidth]{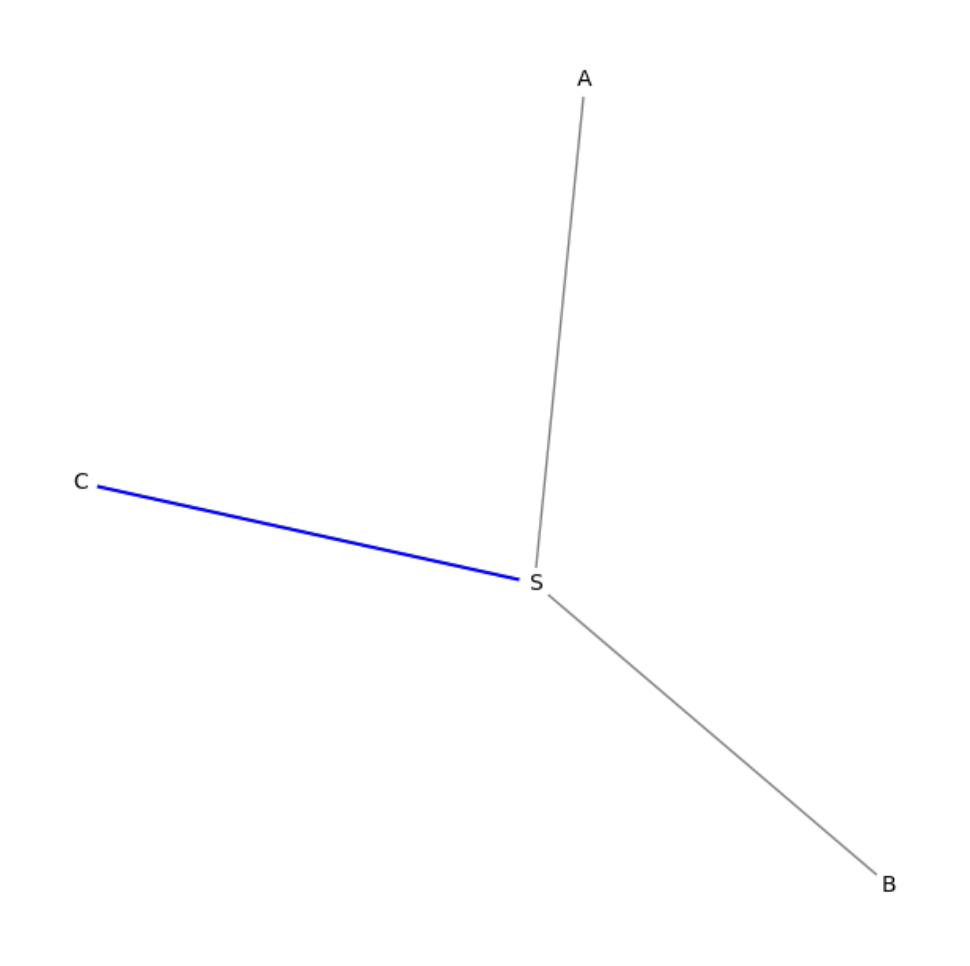}
    \caption{Each graph shows a central node, S, connected by a black edge to each of the nodes: A, B, C. The top-left graph shows the edge connecting node S and node A in blue. The top-right graph shows the edge connecting node S and node B in blue. The lower graph shows the edge connecting node S and node C in blue.}
    \label{fig:adjacency}
\end{figure}

\newpage

The ordering of the state-neighbour dictionary created from the graphs shown in Figure \ref{fig:adjacency} would be: $\mathcal{D} = \left\{ \{C: [S] \}, \{S: [A], [B], [C]\}, \{A: [S]\}, \{B: [S]\} \right\} $. The state-neighbour dictionary forms the basis of the rotation system. The ordered dictionary entries provides a combinatorial embedding for the graph on the plane, by description of the spatial orientation of the states and their corresponding actions via the index of the ordered neighbours of each state. This embedding allows the agent to make decisions based on relative positions rather than absolute coordinates. The relative positions provided by the indexing of the state-neighbour dictionary is shown in the Figure \ref{fig:adjacency}. This example state-neighbour dictionary, state $S$ has three actions $[A], [B], [C]$ which are sorted in a clockwise order. The neighbour ordering translates to a rotation system around the vertex $S$, with edges $S-A$, $S-B$, and $S-C$, ordered clockwise.

The number of available actions for a state depends on the number of neighbours. Most nodes, located in the interior of the graph, have three edge connections and thus three available actions for interior states. For states that have three available actions, the available actions are either \textit{back}, \textit{left}, or \textit{right}. Notice that the action will depend on the relative direction of the previous state. The index of the state-neighbour dictionary then needs to correspond to the action which takes the agent to the neighbour state with the specified relative position to the current state. For a visual example of this this mapping-to-ordered-mapping, let us look at Figure \ref{fig:adjacency}.

The graph $G(\{A, B, C, S \}, \{[A,S], [B,S], [S,C]\})$ in Figure \ref{fig:adjacency}, has the corresponding state-neighbour dictionary $\mathcal{D} = \left\{ \{C: \{S\} \}, \{S: \{A, B, C\}\}, \{A: \{S\}\}, \{B: \{S\}\} \right\}$. This is the result of the order of the states $A$, $B$, and $C$ about state $S$ being alphabetical in a clockwise direction, relative to due north of state $S$. If the agent has just traversed the blue edge in the top-left graph, bringing the agent from state $A$ to state $S$, then the action of \textit{left} should map to the $S-B$ edge that leads to state $B$, whereas \textit{right} would take the $S-C$ edge to neighbouring state $C$ and finally \textit{back} would take blue edge back to previous state $A$. In the top-right graph, traversing the blue edge to state $S$ makes the \textit{left} action the $S-C$ edge to state $C$, the \textit{right} action to state $A$, and the \textit{back} action bringing the agent to previous state $B$. For the lower graph, going \textit{left} leads to state $A$, \textit{right} to state $B$, and \textit{back} to state $C$. Corresponding to the blue edge for the top-left, top-right, and lower graphs in Figure \ref{fig:adjacency}, the state-neighbour dictionary map of state $S$ with the previous state coloured blue is $ \{S: \{\textcolor{blue}{A}, B, C\},  \{A, \textcolor{blue}{B}, C\}, \{A, B,\textcolor{blue}{C}\}\}$, respectively.  
Notice how when current index $i$ refers to a blue coloured state, taking the edge to this neighbour corresponds to the \textit{back} action because the blue state is the other state on the blue edge connection to $S$. Then, with the clockwise ordering, the state next index, $i \equiv 1 \mod 3$, always corresponds to the \textit{left} action and the state back an index, $i \equiv -1 \mod 3$ to the \textit{right}, where $\mod$ denotes the modulo operation.

For a state $s$ with $3$ neighbouring states, the actions associated with this state are $\mathcal{A}(s) = \{back, \; left, \; right\}$. Let previous state, $\varsigma$, be located at the $i$-th index in the state-neighbour dictionary for state $s$. The action $a \in \mathcal{A}(s)$ that corresponds to the $\iota$-th index of the state-neighbour dictionary of state $s$ given previous state $\varsigma$ designated by equation \ref{ref:indicing}.

\begin{equation}
    a  = 
    \begin{cases}
        back & \text{if} \; \iota = (i) \\
        left & \text{if} \;  \iota \equiv (i+ 1) \mod 3 \\
        right & \text{if} \;  \iota \equiv (i- 1) \mod 3 \\
    \end{cases}
    \label{ref:indicing}
\end{equation}

Note that the nodes on the perimeter of the graph have only two connected edges. Refer to Appendix B for a discussion on the actions associated with states with less than and greater than 3 neighbours.

The Q-table, denoted as $Q(s, a)$, is initialised with a value of $1.0$ for each of the state-neighbour pairs obtained from the base graph (i.e. the graph from $\alpha = 0$ perturbation). The base graph is chosen for initialisation because it is on this graph the the agent is first trained.

\subsection{Training}

During the training process, the Q-table is updated iteratively using the Q-learning update rule described in Section 2.1.2. The goal state is the eastern-most (right-most) node on the graph, which sits at the same coordinate location for each graph due to the consistency of the external site placements. In each episode of training, the agent starts from a random state and selects actions following the $\epsilon$-greedy strategy. Taking the selected action leads to a new state and yields a classic reward value of either $1$ when the goal state is reached or $0$ otherwise. The discount factor $\gamma$ is set less than $1$ to give an incentive to minimise the steps taken to reach the goal state.

The modelling choice of the $\epsilon$-greedy strategy was chosen for its simplicity, computational efficiency, and flexible tuning of the exploration-exploitation trade-off. This was chosen instead of the common alternatives defined in the Section 2.1.4, because the $\epsilon$-greedy approach does not require maintaining additional counts of actions or calculating probability distributions over actions, aiding its computationally efficiency. To shift from exploration to exploitation as the agent learns more about the task, the value of $\epsilon_t$ starts with $\epsilon_o = 0.1$ and decays over time by $\epsilon_t = \sqrt{\frac{\epsilon_o}{t}}$ for each step $t$ of the episode.

The action selection process is based on the current graph, which may differ from the graph used to previously construct the Q-table. So, the agent selects the valid actions for the current graph, and due to the consistent action labelling enabled by the rotation system, the state-neighbour pair is typically present in the Q-table. If a state-neighbour pair is not present (e.g. when a node with more than three edges appears in a perturbed graph), the state-neighbour pair is added to the Q-table with a designed initial value of $1.0$ to incentives its exploration. This adaptability is vital for the agent to adjust to new tasks which manifests in a changes in the number of connections of a node in the graphs.

\section{Evaluation}

All agents are first trained on an initial task, given by a base graph of $\alpha=0$ perturbation. The trained Q-table is then updated by further training trained on a similar task which is given by a graph of $\alpha$-perturbation. The agent's performance on the initial task is used to assess the degree of forgetting, using the process outlined in this section.

\subsection{Performance}

To evaluate the learned policy on the initial task, all agents began from the starting state, in this case, the western-most (i.e. left-most) node of the graph. Guided by the trained Q-table, the agents takes actions until it reaches the goal state.

The task (i.e. structure of the graph on which the agent is operating) determines the valid actions an agent can execute. The agent's movements, from the actions taken, utilise the rotation system employed during training. To select an action, each agent calculates the softmax probabilities (as per Definition 3) of the corresponding Q-values for the valid actions of the current state. The Boltzmann exploration was selected for its distributional approach to maintain a balance between exploration and exploitation during the testing phase. This balance was regulated by varying the temperature parameter, $\tau$. Lower values rendered the action selection more deterministic, while higher values promoted a smoother, more exploration-inclined selection distribution. In the experiments, $\tau = 0.1$ was chosen to draw primarily from the learned knowledge, while allowing sufficient exploration to avoid cyclical paths that could prevent reaching the goal state.

\subsection{Forgetting}

The degree of forgetting was measured by the absolute value of the difference in performance after training on a new task. Let the path length to reach the goal state for task $t$ be denoted as $P_t$, and $|\cdot|$ as the absolute value function. Then, $F_t$ can be calculated using Equation \ref{equation:forgetting}.

\begin{equation}
F_t = |P_t^{\text{original}} - P_o^{\text{original}}|
\label{equation:forgetting}
\end{equation}

\subsection{Experiments}

\begin{figure}[!htb]
\centering
\includegraphics[width=0.325\textwidth]{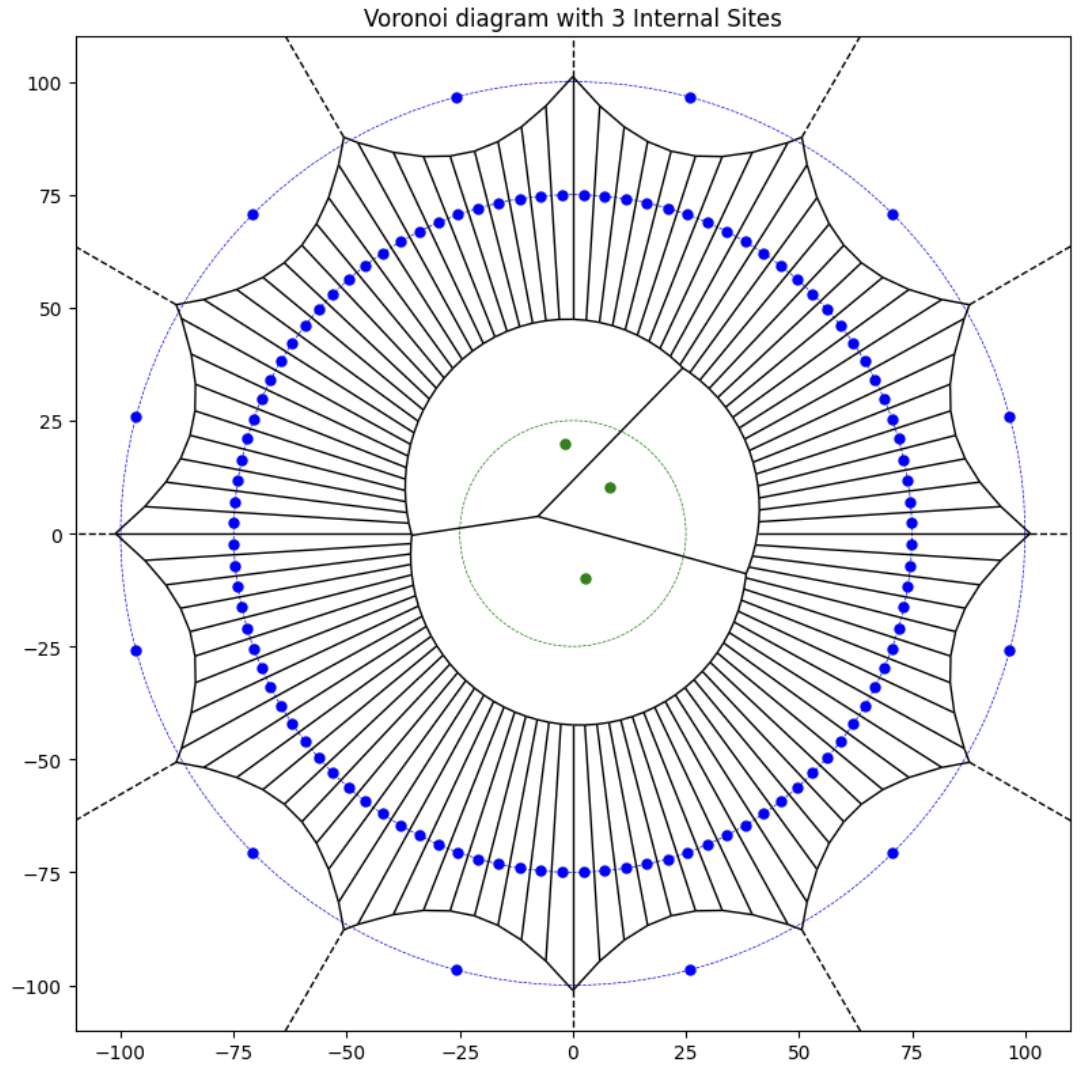} \includegraphics[width=0.325\textwidth]{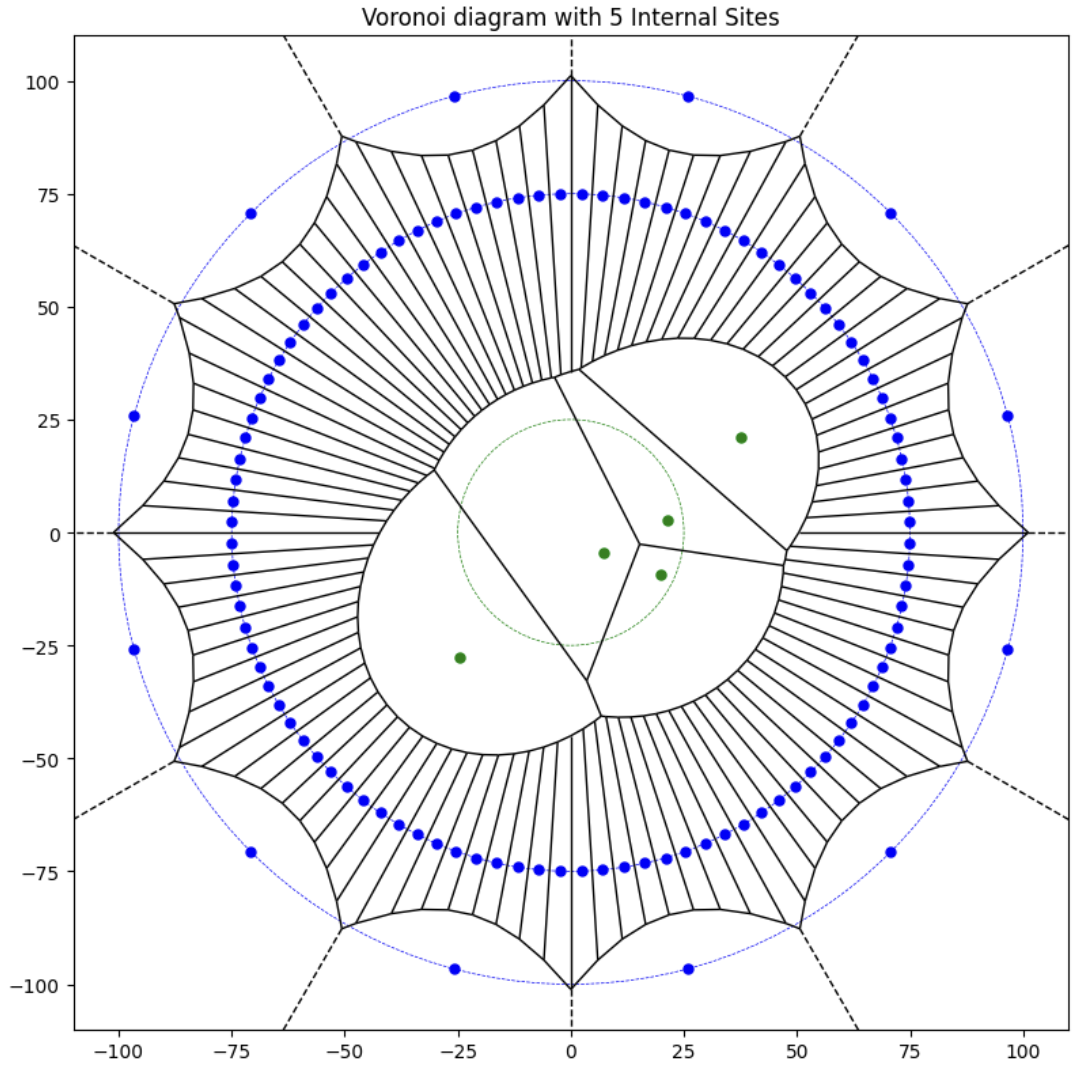}
\includegraphics[width=0.325\textwidth]{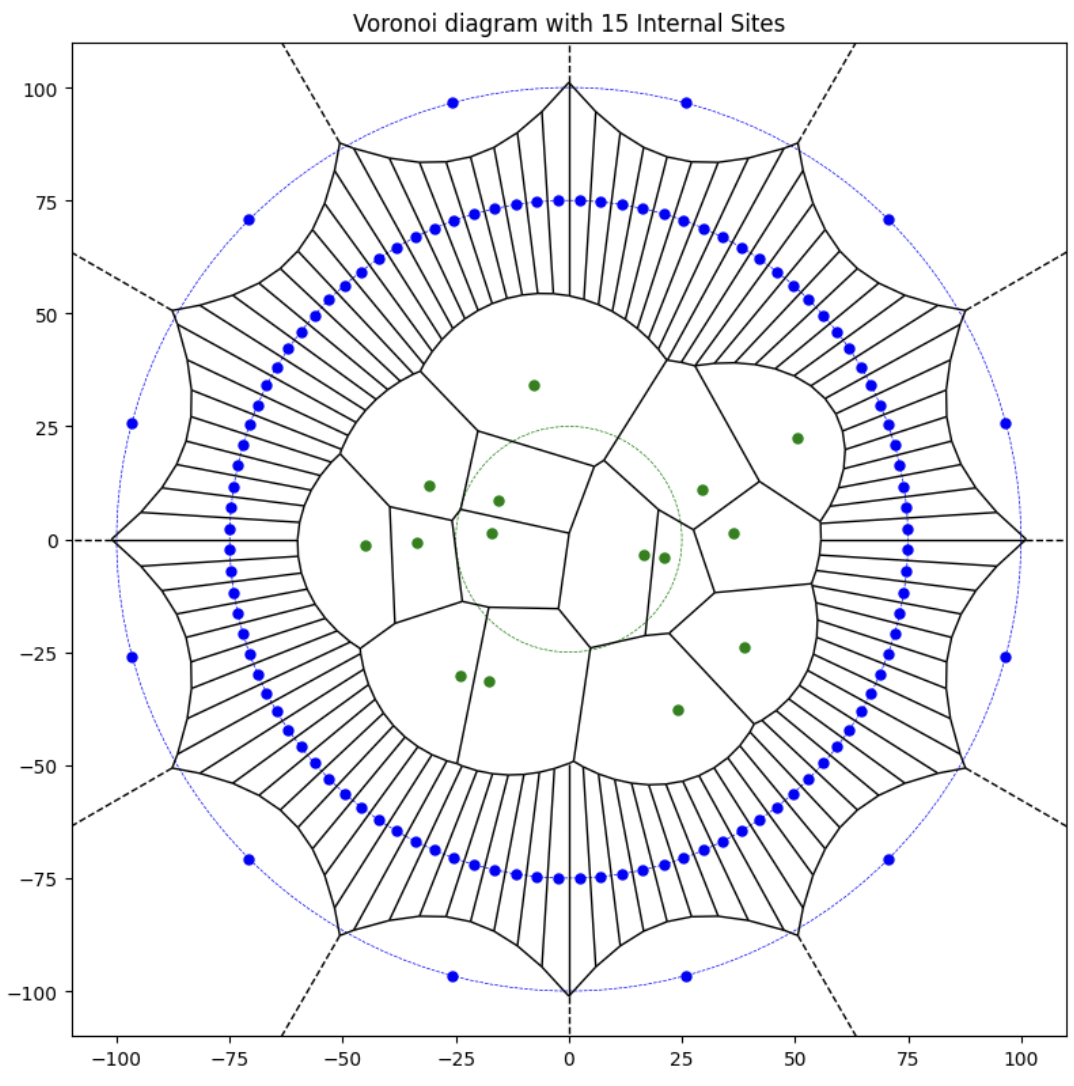}
\scriptsize{(a) \quad \quad \quad \quad \quad \quad \quad \quad  (b) \quad \quad \quad \quad \quad \quad \quad \quad \quad (c)}
\caption{The Voronoi diagram, in black, generated from both the external sites, in blue, and 3 internal sites, in green, is shown in (a). Figure (b) delivers the same visual but with a resulting Voronoi diagram from 5 green, internal sites, and figure (c) displays the diagram with 15 green, internal sites.}
\label{fig:graphs}
\end{figure}

Each trial involved training an agent on a set of graphs derived from Voronoi diagrams, created from sites with incrementally increasing perturbations. One hundred trials were executed on three levels of task complexity, which corresponded to the sizes of the Voronoi diagram graphs generated from different quantities of internal sites. The sizes tested were derived from three, five, and fifteen internal sites, each combined with twelve \textit{outer} external sites and one hundred \textit{inner} external sites, arranged as detailed in Section 4.2.1. Figure \ref{fig:graphs} shows Voronoi diagrams generated from the three quantities of internal sites, and depicts the shape of the resulting diagram from the arrangement of the external sites (as per Section 4.1.1). The site configuration both encourages the agent to pass through the center of the graph, where the controlled internal site manipulation occurs, and to provide a clock-like structure for an interesting start state (western-most node) and goal state selection (eastern-most node) on the graphs from these diagrams. 

Each trial employed a set of graphs derived from Voronoi diagrams with sites perturbed by ten incremental $\alpha$-values from $0$ to $0.25$. Following the initial round of training for $2500$ episodes on the graph from $\alpha = 0$, each agent receives training for another $2500$ episodes on a, respective, perturbed graph. Each agent was evaluated on its performance on the initial task, which involves reaching the eastern-most node from the western-most node on the unperturbed graph. The number of steps taken, the performance, of each agent was recorded for each level of task complexity. To reiterate, the number of steps taken on the original graph (the initial task) serves as the performance metric. The difference between the agent's performance after training on the initial task and the performance after additional training on a different task provides the measure of \textit{forgetting}.

\chapter{Results}

This chapter provides results derived from a hundred experimental trials. Limitations of the study are expressed after presenting and discussing the results.

\section{Experimental Results}

\subsection{Tasks}

\subsubsection{Generated Similar Tasks}

\begin{figure}[!htb]
\centering
\includegraphics[width=1\textwidth]{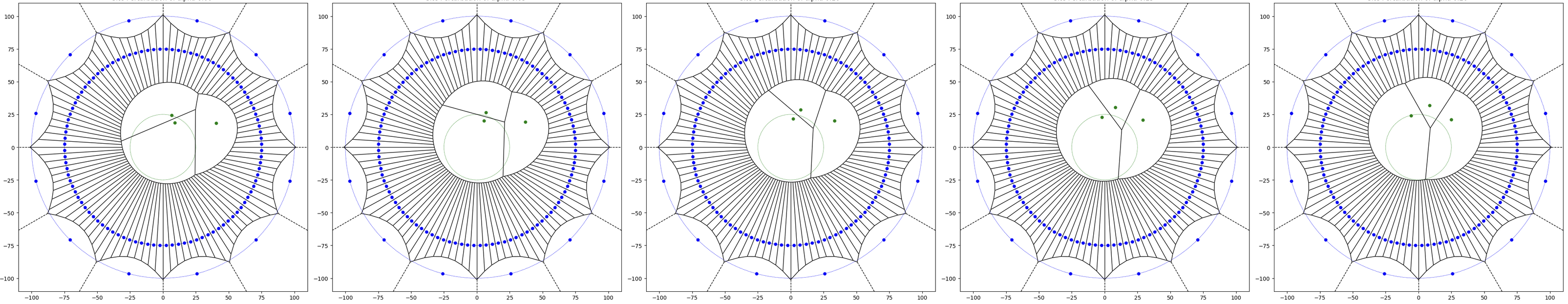}
\small{(a) \quad\quad\quad\quad\quad\quad\quad (b) \quad\quad\quad\quad\quad\quad\quad\quad (c) \quad\quad\quad\quad\quad\quad\quad\quad  (d) \quad\quad\quad\quad\quad\quad\quad (e)}
\caption{\small{Voronoi diagrams from three internal sites with (a) no site perturbation, (b) sites perturbation of $\alpha = 0.05$, (c) sites perturbation of $\alpha = 0.10$, (d) sites perturbation of $\alpha = 0.15$, (e) sites perturbation of $\alpha = 0.20$.}}
\label{fig:vor_3}
\end{figure}

\begin{figure}[!htb]
\centering
\includegraphics[width=1\textwidth]{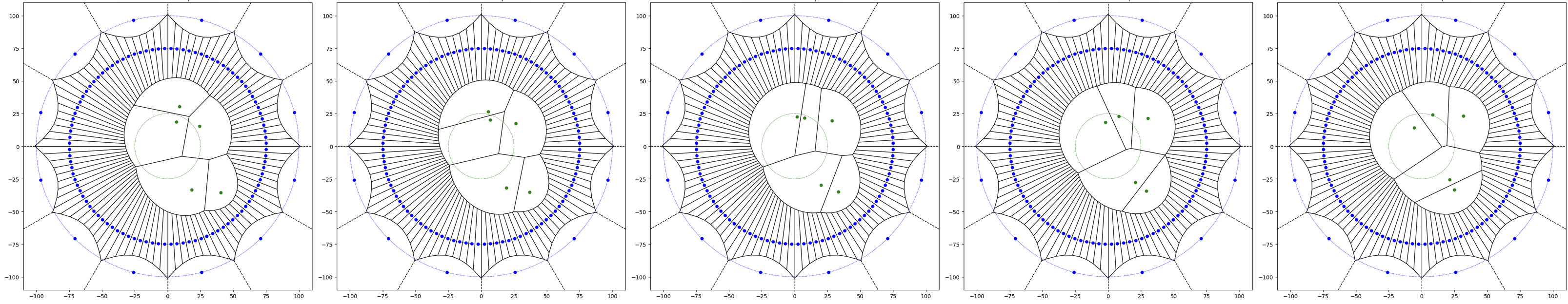}
\small{(a) \quad\quad\quad\quad\quad\quad\quad (b) \quad\quad\quad\quad\quad\quad\quad\quad (c) \quad\quad\quad\quad\quad\quad\quad\quad  (d) \quad\quad\quad\quad\quad\quad\quad (e)}
\caption{Voronoi diagrams from five internal sites with (a) no site perturbation, (b) sites perturbation of $\alpha = 0.05$, (c) sites perturbation of $\alpha = 0.10$, (d) sites perturbation of $\alpha = 0.15$, (e) sites perturbation of $\alpha = 0.20$.}
\label{fig:vor_5}
\end{figure}

\begin{figure}[!htb]
\centering
\includegraphics[width=1\textwidth]{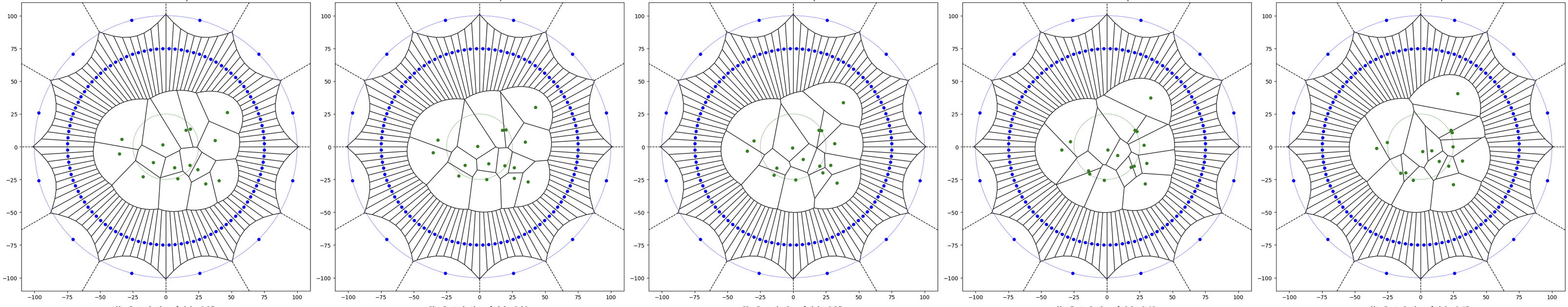}
\small{(a) \quad\quad\quad\quad\quad\quad\quad (b) \quad\quad\quad\quad\quad\quad\quad\quad (c) \quad\quad\quad\quad\quad\quad\quad\quad  (d) \quad\quad\quad\quad\quad\quad\quad (e)}
\caption{Voronoi diagrams from fifteen internal sites with (a) no site perturbation, (b) sites perturbation of $\alpha = 0.05$, (c) sites perturbation of $\alpha = 0.10$, (d) sites perturbation of $\alpha = 0.15$, (e) sites perturbation of $\alpha = 0.20$.}
\label{fig:vor_15}
\end{figure}

Voronoi diagrams were generated for three different sample sizes of internal sites. The perturbations of these sites are displayed in Figures \ref{fig:vor_3}, \ref{fig:vor_5}, and \ref{fig:vor_15}. To avoid an excess of diagrams, every-other $\alpha$-value, from ten increments in the range $[0, 0.25]$, that control the perturbation are displayed in the figures.

\vspace{2.5mm}
\textit{First Task Complexity: 3 Internal Sites}

A set of the Voronoi diagrams generated by 3 internal sites and perturbed by $\alpha$-values are depicted in Figure \ref{fig:vor_3}. It can be seen that as $\alpha$ increases, the edges that form the connections between states can lead to different states, directions, or relative areas. It can also be seen that this perturbation range enables the successful creation of Voronoi diagrams that differ in degrees that maintain relation to the base structure.

The perturbations of just three internal sites offer a difference in tasks that most closely corresponds to the difference in the $\alpha$ perturbations. From the agent's perspective, the perturbations change the states that provide access to the shortcut through the center of the graph. The change in shortcut-access states corresponds to a connection distance generally proportional to the $\alpha$-value of the perturbation. Key for the agent will be remembering how to get to the shortcut-access states of the first graph.

\vspace{2.5mm}
\noindent{\textit{Second Task Complexity: 5 Internal Sites}}

Voronoi diagrams generated with 5 internal sites, perturbed by specified values of $\alpha$, are shown in Figure \ref{fig:vor_5}. From the perspective of the agent, the perturbation of the sites results in a change in the states that provide access to the central shortcut and the states within the shortcut. The changes to these connections predominantly occur in the shortcut-access states. A change in the connections within the shortcut is only seen in Figures \ref{fig:graphs_5}d and \ref{fig:graphs_5}e. It is also observed that the $\alpha$-range employed for the graph changes creates Voronoi diagrams that do not differ to the point of becoming unrelated.

For agents operating on tasks with the size from five internal sites, the connections between the shortcut-access states remain of high importance. However, performing well on the initial task requires remembering the correct connections between a few states within the shortcut, as well as the locations of the states that enable entry to the shortcut through the center.

\vspace{2.5mm}
\noindent{\textit{Third Task Complexity: 15 Internal Sites}}

The most complex tasks were generated using 15 internal sites and the associated Voronoi diagrams, perturbed by $\alpha$, are exhibited in Figure \ref{fig:vor_15}. The increased number of internal sites leads to a corresponding increase in the complexity of the central shortcut. As $\alpha$ increases, the perturbations of site locations lead to significant changes in the connections between the states and the shortcut-access states. Compared to the previous two complexity levels, changes are more profound due to the increased number of connections.

Tasks corresponding to the 15 internal sites involve remembering the larger number of connections to the central shortcut. In addition to the larger number of shortcut-access states, there is an increase in the number of connections within the shortcut itself. The task requires the agent to not only remember how to reach the shortcut but also to navigate through the shortcut to reach the exit. As $\alpha$ increases, the agent needs to learn a new set of connections within the central shortcut that is different from the initial task.

\subsubsection{Difference Measure of Generated Tasks}

\begin{figure}[!htb]
\centering
\includegraphics[width=1\textwidth]{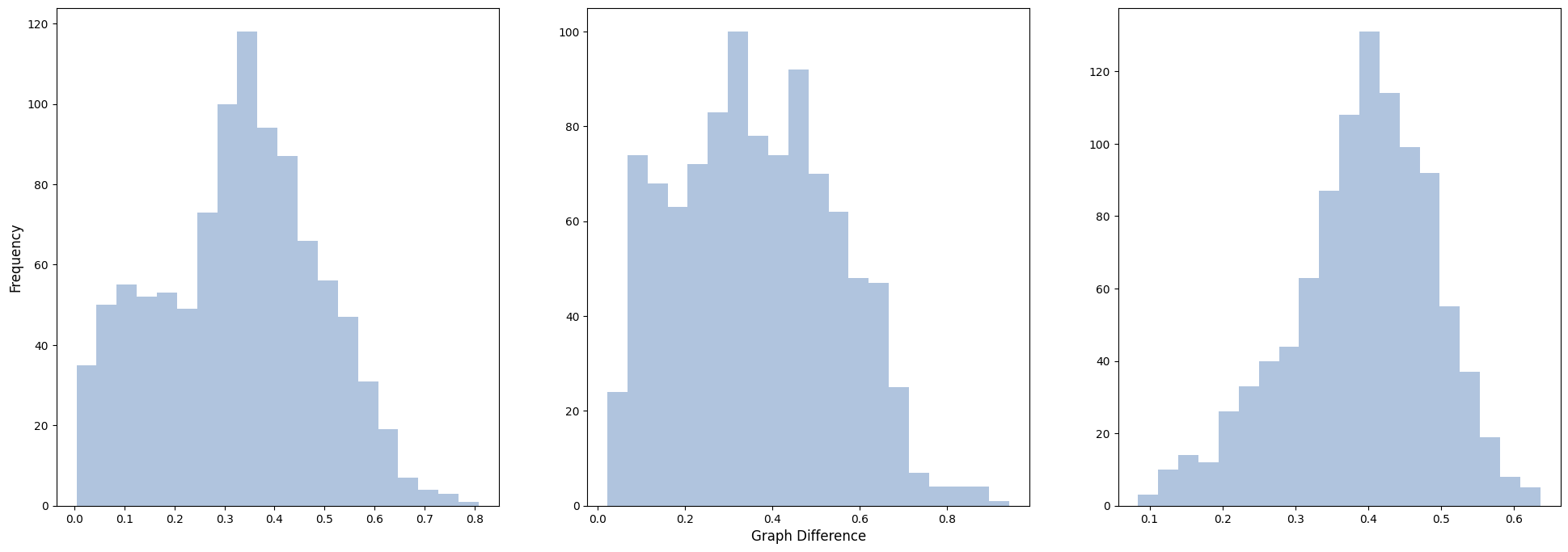}
\small{(a) \quad\quad\quad\quad\quad\quad\quad\quad\quad\quad\quad\quad\quad (b) \quad\quad\quad\quad\quad\quad\quad\quad\quad\quad\quad\quad\quad\quad (c)}
\caption{The histograms depict the number of occurrences of the graph distance measure, giving the task similarity, for all graphs generated in all trials of (a) $M=3$, (b) $M=5$, (c) $M=15$.}
\label{fig:sim_dist}
\end{figure}

The distribution of difference measures between graphs generated from the $\alpha$-perturbations is shown in Figure \ref{fig:sim_dist}. As the x-values progresses to the right on the x-axis, the difference measure between the tasks increases. The bars in the graph represent the average number of occurrences of each similarity measure, plotted over their respective ranges. The results of 3, 5, and 15 internal sites are represented in Figure \ref{fig:sim_dist}a, \ref{fig:sim_dist}b, and \ref{fig:sim_dist}c respectively.

A bar height of 100 represents the occurrence of a specific measure at a rate of once per trial. The histograms indicate that for all three sets of internal sites, the intermediate similarity measures occur about once every trial. This observation simply parallels the central limit theorem, however such a severe lack of certain measures is an unexpected observation and limitation. The paucity of graphs with the highest difference measure is significant and is factored into the interpretations throughout this chapter. The disproportionate representation of intermediate similarity measures and the relative scarcity of most-similar graph measures in Figure \ref{fig:sim_dist}c are also noteworthy.

\subsection{Performance}

\begin{figure}[!hbt]
\centering
\includegraphics[width=0.49\textwidth]{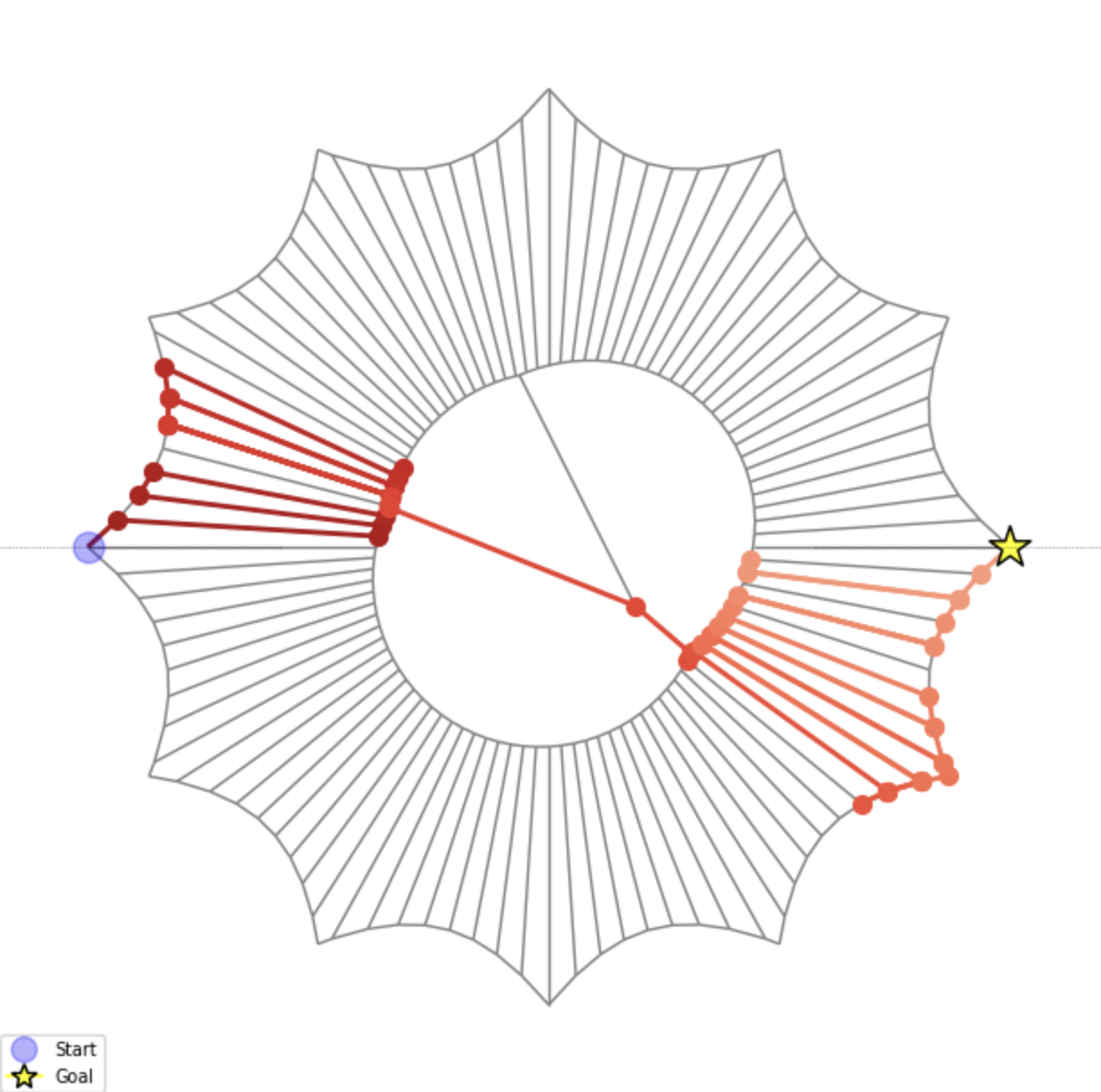} \includegraphics[width=0.49\textwidth]{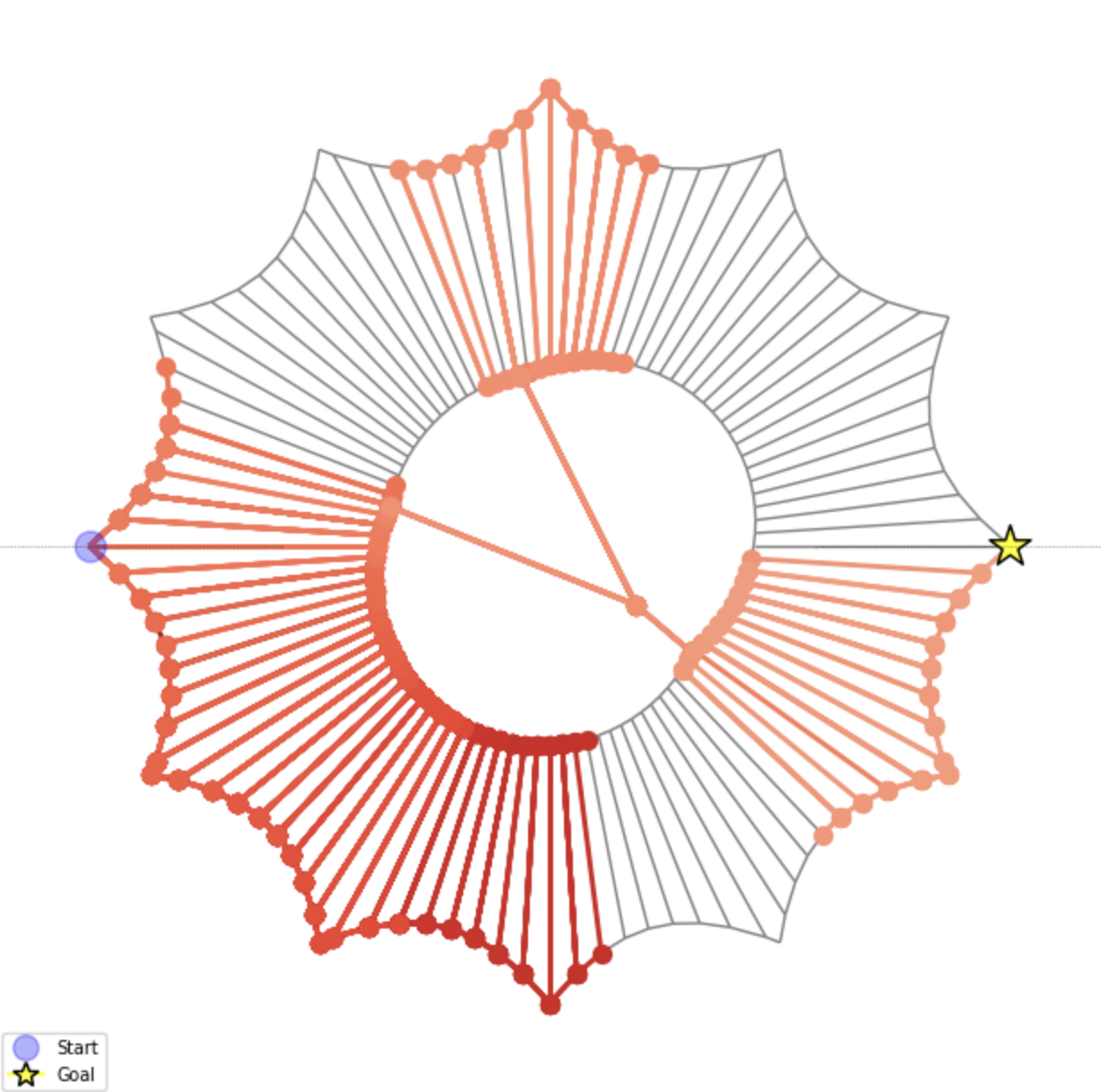}
\small{(a) \quad \quad \quad \quad \quad \quad \quad \quad \quad \quad \quad \quad \quad \quad \quad \quad \quad \quad \quad \quad \quad (b)}
\caption{The graphs are shown in grey. A blue circle marks the start state and a gold star marks the goal state. The path is shown in red with a hue that lightens as the path progresses. Figures (a) and (b) illustrate two performances, respectively, after additional training on different tasks.}
\label{fig:graphs_3}
\end{figure}

\begin{figure}[!hbt]
\centering
\includegraphics[width=0.49\textwidth]{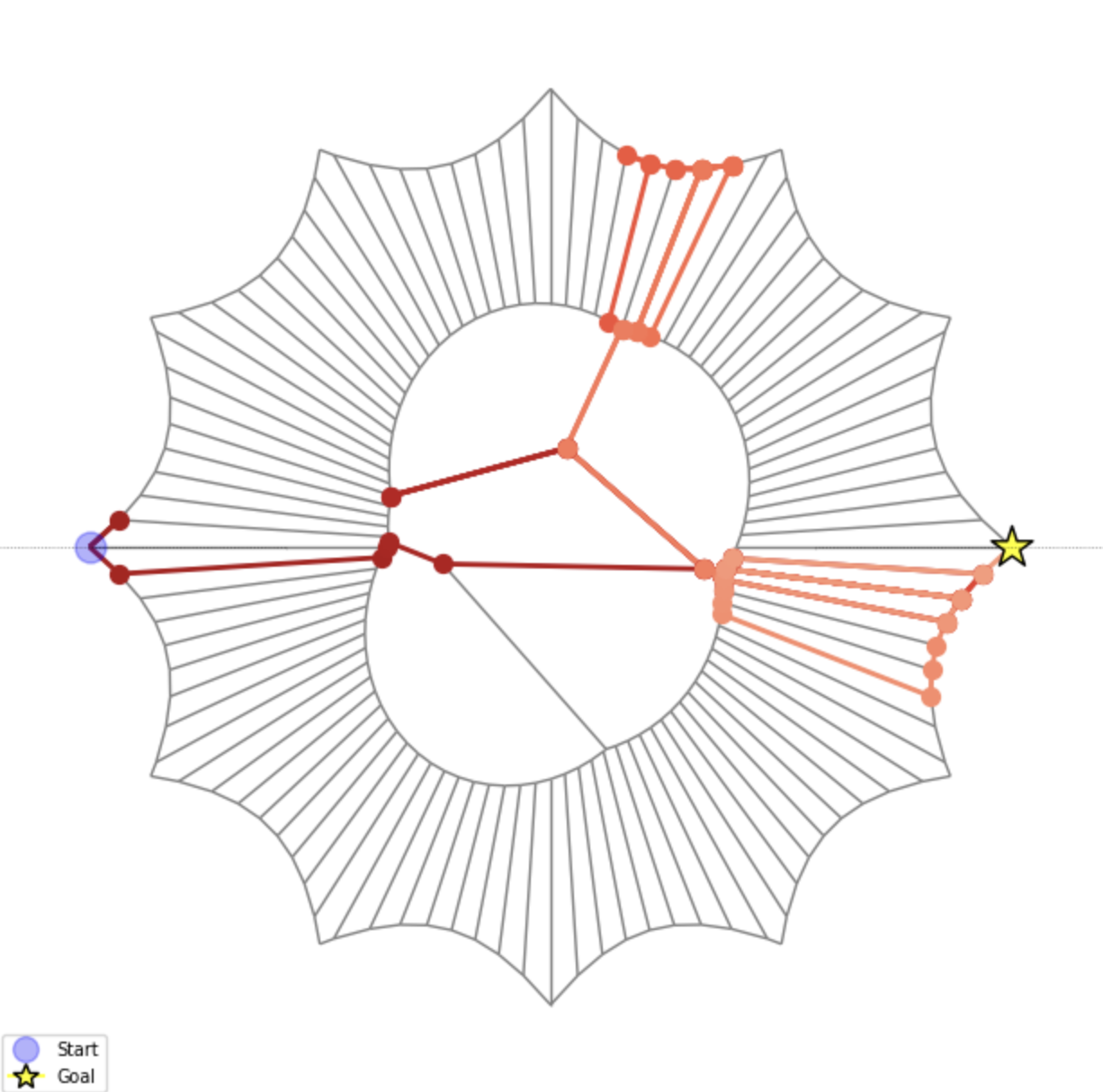} \includegraphics[width=0.49\textwidth]{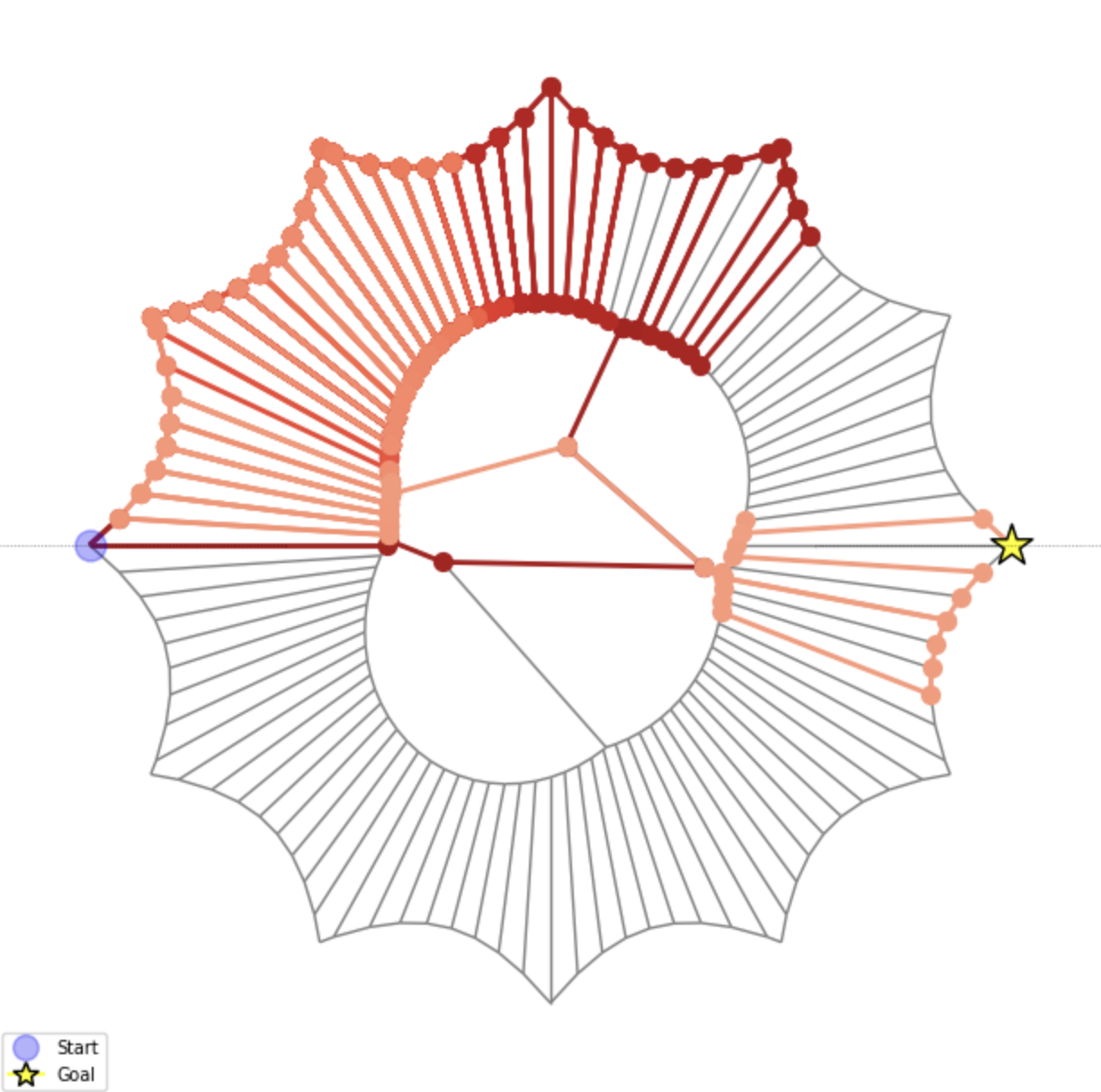}
\small{(a) \quad \quad \quad \quad \quad \quad \quad \quad \quad \quad \quad \quad \quad \quad \quad \quad \quad \quad \quad \quad \quad (b)}
\caption{The graphs are shown in grey. A blue circle marks the start state and a gold start marks the goal state. The path is shown in red with a hue that lightens as the path progresses. Figures (a) and (b) illustrate two performances, respectively, after additional training on different tasks.}
\label{fig:graphs_5}
\end{figure}

\begin{figure}[!hbt]
\centering
\includegraphics[width=0.49\textwidth]{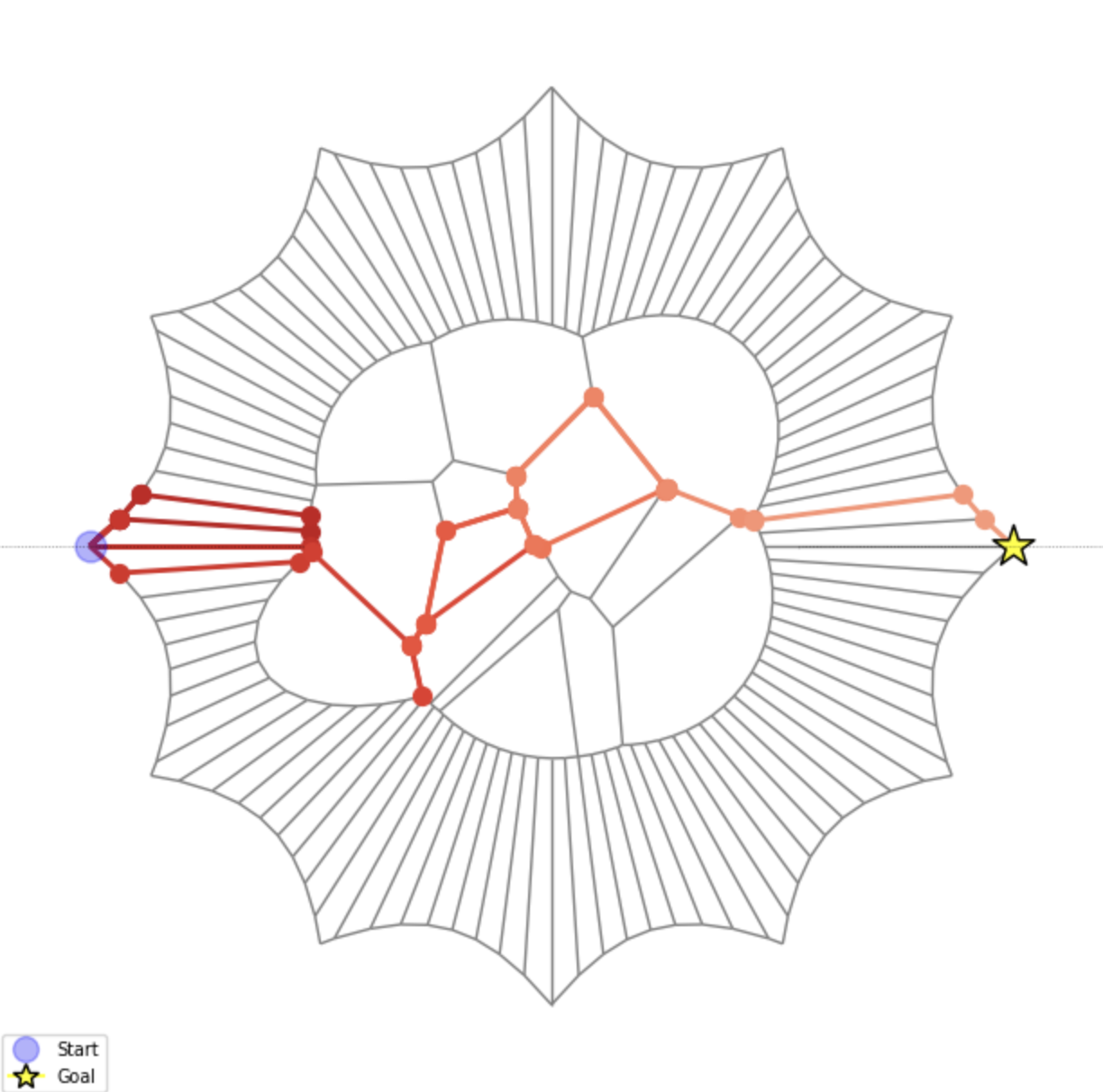} \includegraphics[width=0.49\textwidth]{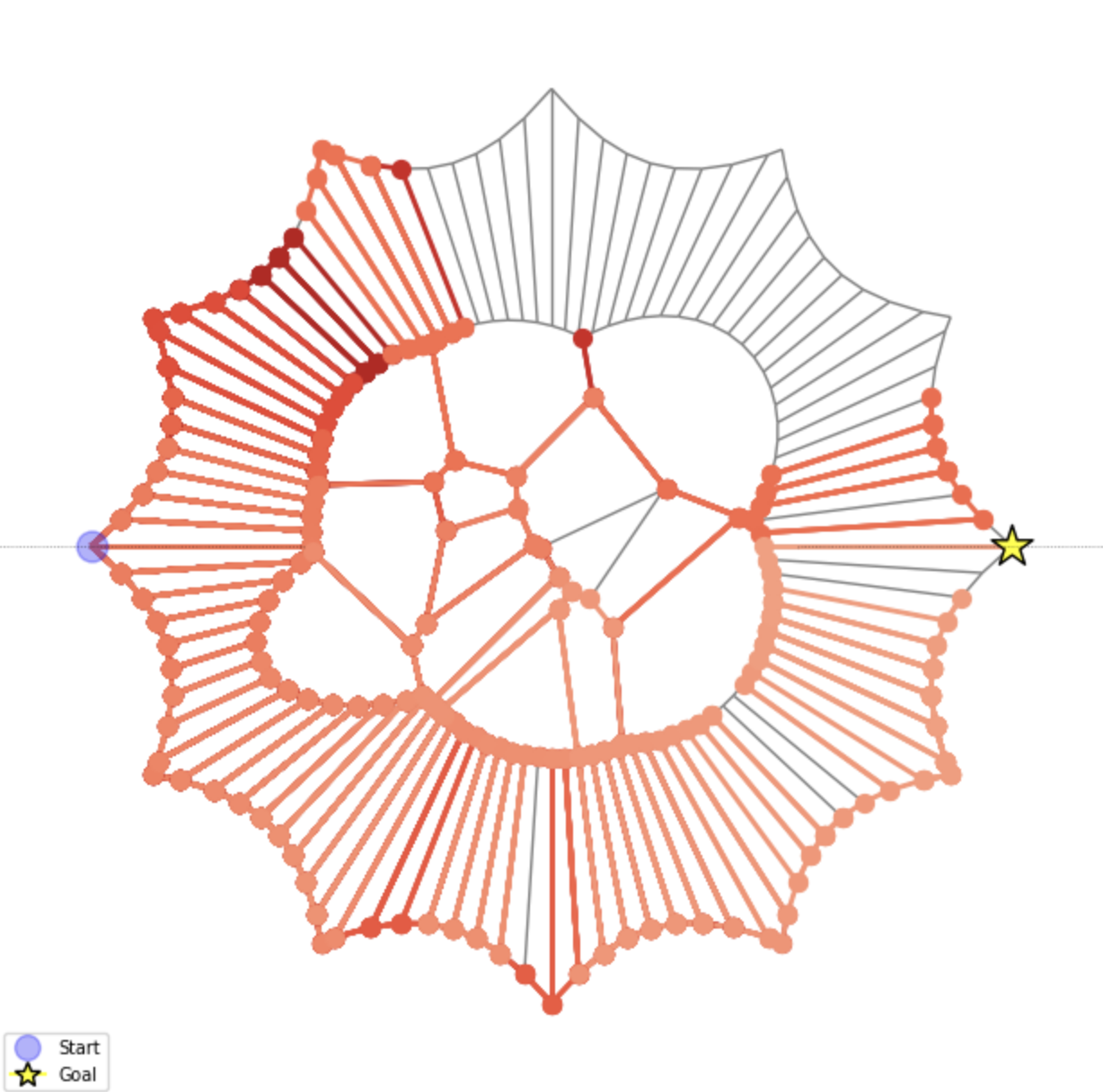}
\small{(a) \quad \quad \quad \quad \quad \quad \quad \quad \quad \quad \quad \quad \quad \quad \quad \quad \quad \quad \quad \quad \quad (b)}
\caption{The graphs are shown in grey. A blue circle marks the start state and a gold start marks the goal state. The path is shown in red with a hue that lightens as the path progresses. Figures (a) and (b) illustrate two performances, respectively, after additional training on different tasks.}
\label{fig:graphs_15}
\end{figure}

Agent performances on the initial task, given by the unperturbed graph, are plotted in Figures \ref{fig:graphs_3}, \ref{fig:graphs_5}, and \ref{fig:graphs_15}. The figures are presented in increasing order of task complexity. Each graph within the figures is a base graph ($\alpha = 0$) generated during a trial. The agent's starting state is denoted by a blue circle, while the goal state is signified by a gold star. The path traversed by the agent is marked in red, with the hue lightening as the agent's steps progress, providing insight into the sequence in which the graph was explored. Figures \ref{fig:graphs_3} - \ref{fig:graphs_15} illustrate a comparison of agent performances: a satisfactory instance in (a) versus a severely suboptimal instance in (b).

Effective navigation of the smallest task size requires learning to reach the nearest state that allows access to the shortcut through the center of the graph. This behaviour appears to have been learned by the agents, as evident in both Figure \ref{fig:graphs_3}a and \ref{fig:graphs_3}b, where the states around each shortcut-access state are evident.

On the intermediate task size, the performance of the agent in Figure \ref{fig:graphs_5}a shows a better memory of the location of the shortcut-access states but operates in the same sub-optimal manner on the navigation of the shortcut states themselves as the agent in Figure \ref{fig:graphs_5}b. The remembered location, or learning of the location, of these shortcut-access states by the agents remains a key distinguishing factor between good and poor performance. However, the difference between good and optimal performance also hinges on the knowledge of the best path in the shortcut itself.

The high complexity of the shortcut provided by the graphs of Figure \ref{fig:graphs_15} underscores the importance of knowledge about the shortcut itself when it comes to the largest task size. Figure \ref{fig:graphs_15}a illustrates a case where an agent circulates around the shortcut-access state several times before effectively, albeit imperfectly, traversing the cut-through. Conversely, the performance showcased in Figure \ref{fig:graphs_15}b demonstrates how poor execution in the shortcut can result in additional steps from disorientation, as well as increased potential to exit the central shortcut area. The need to locate the shortcut-access states and navigate the states in the shortcut becomes most evenly divided for agents operating on the largest task size.

\subsection{Relationship between Task Similarity and Forgetting}

\begin{figure}[!htb]
\centering
\small{ (a) \quad\quad\quad\quad\quad\quad\quad\quad\quad\quad\quad\quad\quad (b) \quad\quad\quad\quad\quad\quad\quad\quad\quad\quad\quad\quad\quad (c)}
\includegraphics[width=1\textwidth]{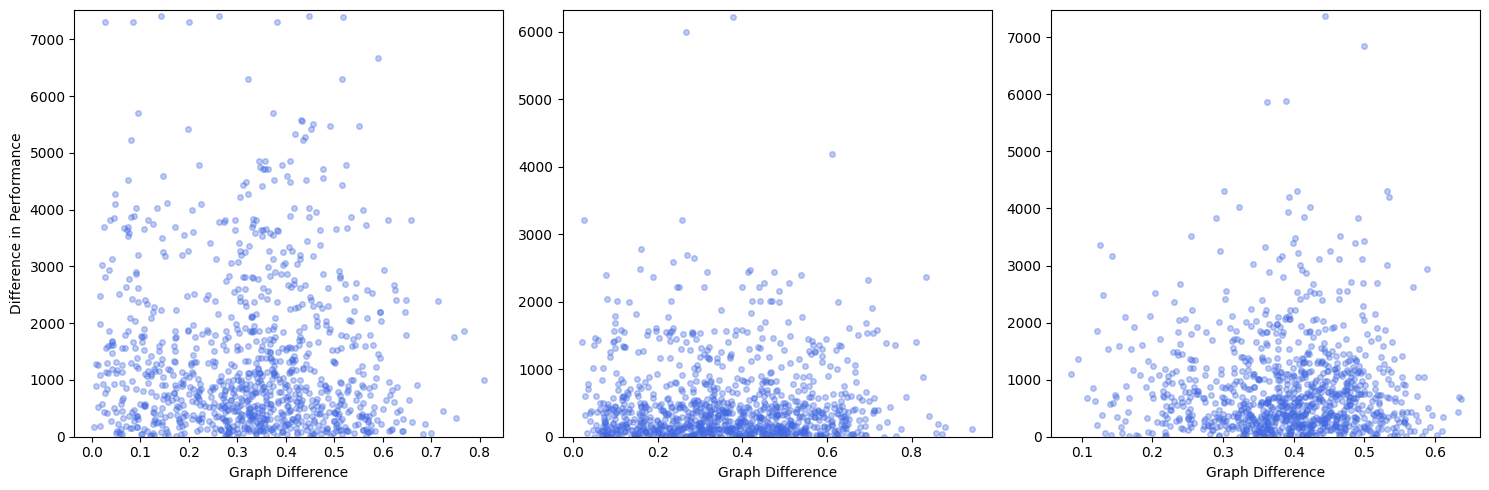}
\caption{The blue markers represent the difference in performance on the y-axis given the measure of difference between graphs on the x-axis. Graphs (a), (b), and (c) correspond with the trials of the first, second, and third graph sizes, respectively.}
\label{fig:raw_relationship}
\end{figure}

\begin{figure}[!htb]
\centering
\small{ (a) \quad\quad\quad\quad\quad\quad\quad\quad\quad\quad\quad\quad\quad (b) \quad\quad\quad\quad\quad\quad\quad\quad\quad\quad\quad\quad\quad (c)}
\includegraphics[width=1\textwidth]{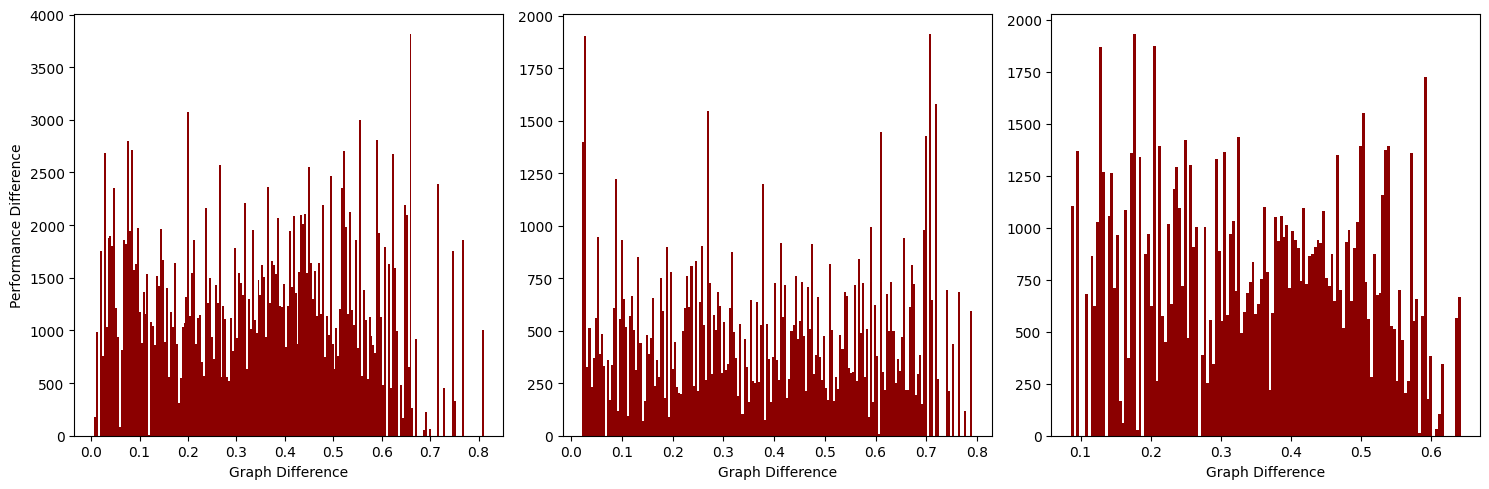}
\caption{The bar heights represent the average difference in steps and the bar widths represent the graph difference ranges. The span of the graph differences is partitioned into 500 ranges. Graphs (a), (b), and (c) correspond with the trials of the first, second, and third graph sizes, respectively.}
\label{fig:relationship_200}
\end{figure}

\begin{figure}[!hbt]
\centering
\small{ (a) \quad\quad\quad\quad\quad\quad\quad\quad\quad\quad\quad\quad\quad (b) \quad\quad\quad\quad\quad\quad\quad\quad\quad\quad\quad\quad\quad (c)}
\includegraphics[width=1\textwidth]{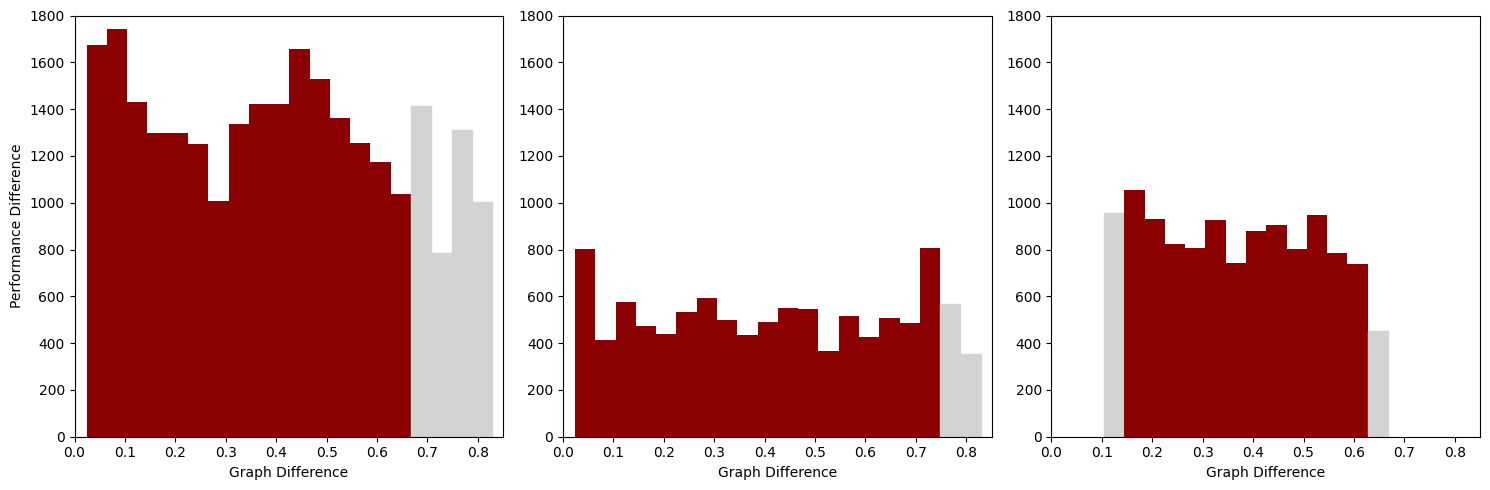}
\caption{The bar heights represent the average difference in steps and the bar widths represent the graph difference ranges. The span of the graph differences is partitioned into 20 ranges. The bars are grey when less than ten samples are present. Graphs (a), (b), and (c) correspond with the trials of the first, second, and third graph sizes, respectively.}
\label{fig:relationship_20}
\end{figure}

Figures \ref{fig:raw_relationship}, \ref{fig:relationship_200}, and \ref{fig:relationship_20} depict forgetting in terms of task similarity. In Figure \ref{fig:raw_relationship}, each plot shows the amount forgotten versus the similarity of the new task of a hundred trial for each of the task complexity levels.

The Figure \ref{fig:raw_relationship} displays a high variability in the differences in performance post-training on the new task. Across the task sizes, the forgetting measures appear to be concentrated in the lower amounts and then sees increases into performance differences of $7000$. Quantities of forgetting in the thousands indicates a considerable policy difference after training on the new task. Each task size appears to present a spectrum of forgetting severity, accompanied by instances of catastrophic forgetting.

Figure \ref{fig:raw_relationship}a shows that the smallest task size has a considerably greater forgetting variance than the larger task sizes depicted in Figure \ref{fig:raw_relationship}b and \ref{fig:raw_relationship}c. This could be contributed to the importance of locating the shortcut-accessing states for agents operating on graphs of this size, as discussed in Section 5.1.2. If the agent struggles to reach the shortcut through the graph, it results in bounces around the outer states, considerably lengthening the steps of the path. The impact of task similarity on the amount forgotten is not explicit from Figure \ref{fig:raw_relationship}.

Figure \ref{fig:relationship_200} helps to see past some of the noise by illustrating the average forgetting measures over small task difference intervals. The histograms have two hundred bins covering the range of task differences. For tasks of the first complexity, as shown in Figure \ref{fig:relationship_200}a, significant forgetting fluctuations span the range of task similarities. Notice that the range of performance differences on the y-axis is double that for this task size than for the two larger ones. The fluctuations into catastrophic forgetting are much more frequent than the other two complexities and the overall measure much higher. Tasks of the second complexity, depicted in Figure \ref{fig:relationship_200}b, appear to have a more uniform distribution of forgetting with higher fluctuations at both ends of the similarity scale. Figure \ref{fig:relationship_200}c shows high fluctuations in forgetting at the extremes of the task similarity scale. It is important to note, however, that the ranges with increased fluctuations directly correspond to severely underrepresented task similarity measures, as depicted in Figure \ref{fig:sim_dist}c.

The histograms in Figure \ref{fig:relationship_20} further generalise by averaging forgetting over just 20 similarity intervals. Figure \ref{fig:relationship_20}a, \ref{fig:relationship_20}b, and \ref{fig:relationship_20}c displays the relationship over the full range of graph distance measures for each size task. The differences in the distribution of graph difference generated for the separate task sizes is starkly evident. Figure \ref{fig:relationship_20}c does not contain the most-similar tasks or tasks of a difference as high as those in \ref{fig:relationship_20}a and \ref{fig:relationship_20}b, as observed in Figure \ref{fig:similarity} in the 5.1.1 Task section.

In the figures that contain highly similar tasks, \ref{fig:relationship_20}a and \ref{fig:relationship_20}b, the rate of forgetting for the most-similar tasks is noticeably high for both participating task sizes. As tasks increase in difference from the highest similarity, a drop in the average rate of forgetting is seen in Figures \ref{fig:relationship_20}a and \ref{fig:relationship_20}b. Forgetting decreases significantly, to about half of the most-similar tasks, past those least-different tasks. Tasks that exhibit intermediate levels of similarity also seem to present fewer spikes in catastrophic forgetting, as shown in Figure \ref{fig:relationship_200}b and \ref{fig:relationship_200}c. The patterns observed in the plots of Figure \ref{fig:raw_relationship} suggest that this trend may be the result of variability and instances of catastrophic forgetting, rather than a threshold beyond which the phenomenon occurs. 

As the tasks increase in difference from mid-similarity, the average forgetting in Figures \ref{fig:relationship_20}a and \ref{fig:relationship_20}b see another increase. Across task complexities, however, the increases come at different rates and at inconsistent dissimilarity measures. Despite the limited task difference range present, the even distribution of forgetting in Figure \ref{fig:relationship_20}c shows no effect of task similarity on forgetting. The high inconsistency between all three task complexities indicates that task similarity and task complexity are not mutually independent in their impact on forgetting.

This study does not find evidence of statistical significance that task similarity independently has a discernible effect on forgetting in continual RL. The study suggests that task similarity and task complexity are not independently affecting forgetting, but rather, are interdependent in their impact. Given the high degree of variability of forgetting measures and the uneven distribution of graph differences, the relationship between these variables remains unclear. Further research is warranted to gain a comprehensive understanding of the potential interplay between task similarity on catastrophic forgetting.

\section{Limitations}

A more comprehensive and direct measure for the difference between graph structures would correspond more highly to the amount of difference the agent is believed to interpret between tasks. The inability to utilise the graph edit distance, which directly quantifies the structural transformation cost between two graphs, posed a notable constraint. This metric offers a more direct and capable measure for the similarity between the structural-graph-based tasks than the indirect metric employed. Additionally, more robust results could be achieved if there were a more even distribution of graph differences were present.

Another constraint relates to the precision of computation during the formation of the Voronoi diagrams from the sites. The limited precision of the software was observed through the instances of nodes with five edges during the trials. This phenomenon occurred in the largest graphs, despite its high statistical improbability. Such an anomaly can be attributed to precision limitations on the intersection of the Voronoi polygon edges, where closely placed vertices could be considered identical.

While the research is able to provide insight into this previously unquantified relationship, the results may only be indicative of trends for tabular RL in deterministic graphical settings.

\chapter{Conclusions and Future Directions}

This research explored the relationship between task similarity and forgetting in a continual reinforcement learning. The experimental findings uncovered a complex interaction between these variables, with substantial variability in forgetting measures across task complexities and similarities. 

A spectrum of forgetting severity was observed, showing significant fluctuations across different degrees of task similarity and complexity. These fluctuations, sometimes reaching catastrophic measures, appeared to be uninfluenced by task similarity in isolation. On average, high degrees of forgetting were observed for tasks that were either highly similar or highly dissimilar. These observations were accompanied with the observations of an uneven distribution of graph similarities from the related graphs that were generated. Across the levels of task complexity, the absence of a clear relationship could be attributed to the high variability and uneven distribution of graph differences. 

Interestingly, the results are suggestive that the impact of forgetting may be contingent upon a combined influence of task similarity and task complexity. However, the study did not uncover statistically significant evidence to affirm that task similarity has a measurable effect on forgetting in a continual reinforcement learning setting.

To begin investigating approaches that improve generalisation, it may prove useful to evaluating the ramifications of different modelling hyperparameters, examining a wider spectrum of task complexity, and modifying the objective of the reinforcement learning agent. If the objective were to minimise the total distance of the path, as opposed to the number of steps taken, the perturbation value that generates the related tasks could also serve as a measure of task similarity.

Diving deeper into diverse sequential training scenarios offers a promising trajectory for future studies. Consider the scenario of training an agent back on the initial task post-training on the new task, or having an agent sequentially tackle tasks with increasing differences. Such approaches could shed new light on how different tasks influence forgetting. Additionally, performance evaluations during intermediate stages of training on the subsequent task could provide insights into the evolving performance on the original task.

Employing deep reinforcement learning methods, starting with deep Q-learning, provides opportunities to explore more complex, high-dimensional tasks. Such exploration could offer deeper insights into the effects of task similarity on forgetting, although these might not be as easily interpretable. The recent findings of the effect of task similarity on forgetting in deep learning networks observed that the greatest forgetting occurs with intermediate task similarity \cite{Lee2021ContinualSimilarity, Ramasesh2020AnatomySemantics}. Whether this observation holds true in the reinforcement learning setting may contribute to an understanding of this relationship across fields.

Ultimately, while this research casts light on a previously unquantified relationship in reinforcement learning, the effect of task similarity on forgetting remains largely undefined, warranting further exploration. Intriguing directions for future inquiry and a growing academic interest underscore the significance and potential of this topic in shaping future research.

\addcontentsline{toc}{chapter}{Appendices}

\appendix
\chapter{Modelling Parameter}

\begin{table}[h]
\centering
\begin{tabular}{|c|c|l|}
\hline
Parameter & Value & Definition \\
\hline
\hline
$R$ & $100$ & Radius of outer concentric circle upon which \\
& & the \textit{outer} external sites are dispersed \\
$r$ & $75$ & Radius of inner concentric circle upon which \\
& & the \textit{inner} external sites are dispersed \\
$|N_{outer}|$ & $12$ & Number of \textit{outer} external sites \\
$|N_{inner}|$ & $100$ & Number of \textit{inner} external sites \\
$|M|$ & $3, 5, 15$ & Number of internal sites \\
& & \\

$\alpha$ & $\{0.025, 0.05, \ldots, 0.25 \}$ & Perturbation values for Voronoi diagrams \\

& & \\
episodes & $2500$ & Number of training episodes for each training  \\
$Q(s, a)$ & $1.0$ & Initial Q-value for each state-action pair \\
learning rate & $0.1$ & Q-learning learning rate for Q-table updates \\

& & \\
$\gamma$ & $0.9$ & Discount factor \\
$\epsilon_o$ & $0.1$ & Initial epsilon for $\epsilon$-greedy strategy \\
$\epsilon_{2500}$ & $0.006$ & Final epsilon for $\epsilon$-greedy strategy  \\
\hline
\hline
\end{tabular}
\caption{A summary of model parameters and hyperparameters.}
\end{table}

\label{appendixlabel1}

\chapter{State Actions for Unusual Size Action Sets}
\label{appendixlabel2}

The action-edge mappings correspond to the number of connections to neighbouring states, of which there are not always three. When states have more than three neighbours, a relatively common occurrence with the slow shifting of the sites and the limited precision of the calculation software, additional action options need to be available. Constant for all graphs due to the identical outer cicular site placements, the outer-most states from the center of the graph have only two neighbours. A modification of Equation 4.6, is used for and state $s$ the number of neighbouring state, $|\mathcal{E}(s)|$, different for the encountered amounts of 2, 4, and 5 during training. The actions given are:
\begin{itemize}[noitemsep]
    \item $\mathcal{A}_2(s) = \{back, \; forward \}$ for $|\mathcal{E}(s)|=2$;
    \item $\mathcal{A}_4(s) = \{back, \; left, \; forward, \; right \}$ for $|\mathcal{E}(s)|=4$;
    \item $\mathcal{A}_5(s) = \{back, \; left, \; forward-left, \; forward-right, \; right \}$ for $|\mathcal{E}(s)|=5$
\end{itemize}

To preserve the order, let $A_2 = [back, \; forward]$, $A_4 = [back, \; left, \; forward, \; right]$, and $A_5 = [back, \; left, \; forward-left, \; forward-right, \; right]$. Let previous state, $\varsigma$, be located at the $i$-th index in the state-neighbour dictionary of state $s$. The action $a \in \mathcal{A}(s)$ that corresponds to the $\iota$-th index of the state-neighbour dictionary of state $s$ given previous state $\varsigma$ designated by equation \ref{ref:indicing}.

\begin{equation}
    a  = A_{|\mathcal{E}(s)|}(\iota) \; \text{for} \; \iota \equiv (i+ \eta) \mod |\mathcal{E}(s)|, \; \; \;  \forall \; \eta = 0, 1, \dots, (|\mathcal{E}(s)| - 1)
    \label{ref:indicing_others}
\end{equation}

\printbibliography[title = {Bibliography}]

\end{document}